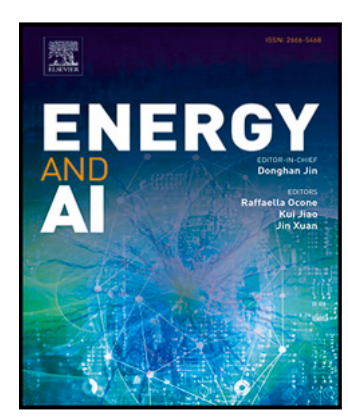

# A flow-based model for conditional and probabilistic electricity consumption profile generation ☆

Weijie Xia [a], Chenguang Wang [b], Peter Palensky [a], Pedro P. Vergara [a,*]

[a] Intelligent Electrical Power Grids (IEPG) Group, Delft University of Technology, Delft, 2628 CD, Netherlands
[b] Alliander N.V, Arnhem, 6812 AH, Netherlands

## HIGHLIGHTS

- The proposed model enables high-quality generation under continuous conditions.
- Statistical models capture global accuracy but struggle with complex distributions.
- Deep generative models capture temporal complexity but miss global accuracy.
- The proposed model blends the advantages of deep generative and statistical methods.

## GRAPHICAL ABSTRACT

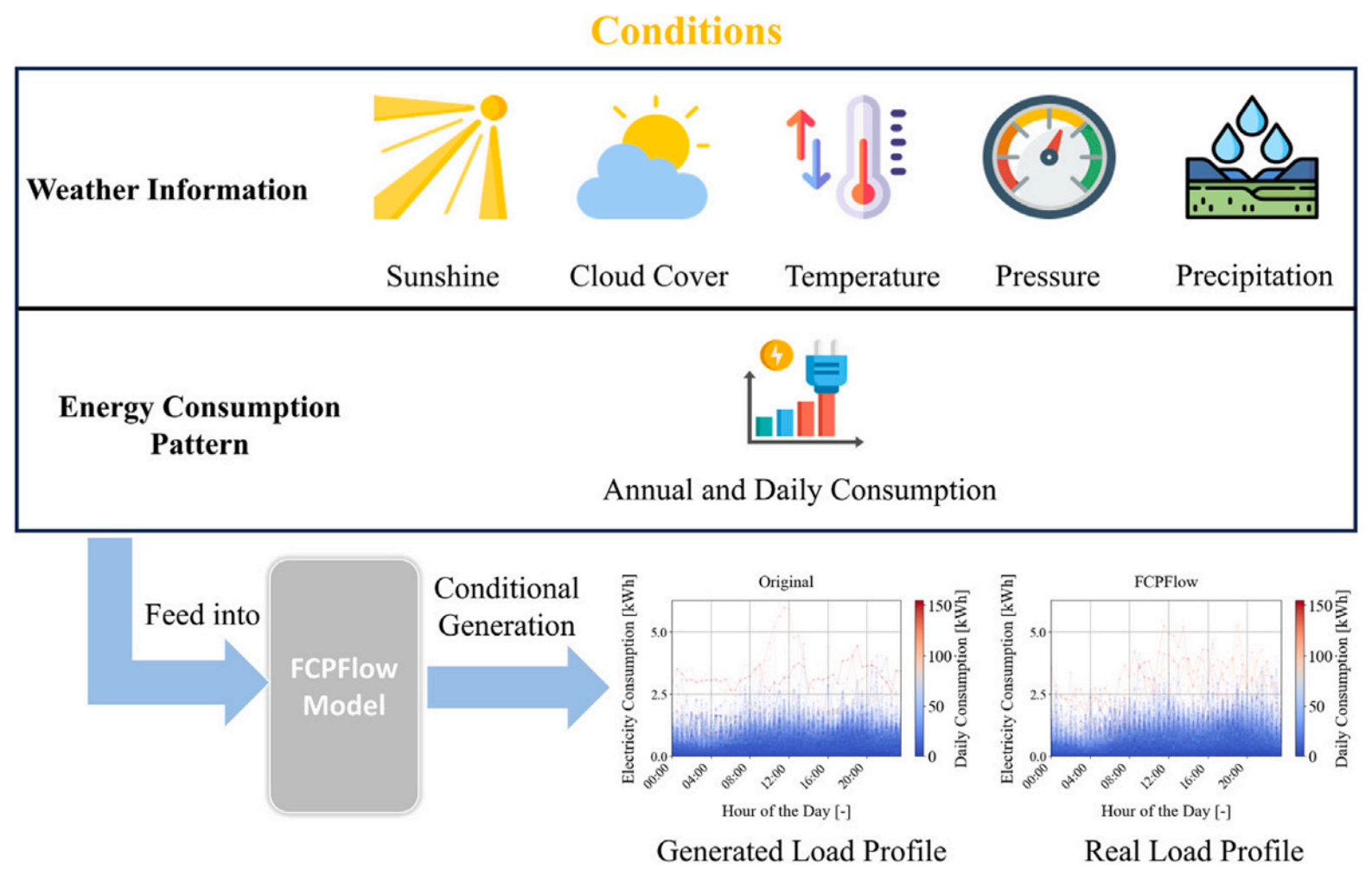

## ARTICLE INFO

## ABSTRACT

Residential Load Profile (RLP) generation is critical for the operation and planning of distribution networks, especially as diverse low-carbon technologies (e.g., photovoltaic and electric vehicles) are increasingly adopted. This paper introduces a novel flow-based generative model, termed Full Convolutional Profile Flow (FCPFlow), uniquely designed for conditional and unconditional RLP generation. By introducing two new layers – the invertible linear layer and the invertible normalization layer – the proposed FCPFlow architecture shows three main advantages compared to traditional statistical and contemporary deep generative models: (1) it is well-suited for RLP generation under continuous conditions, such as varying weather and annual electricity consumption, (2) it demonstrates superior scalability in different datasets compared to traditional statistical models, and (3) it also demonstrates better modeling capabilities in capturing the complex correlation of RLPs compared with deep generative models.

☆ This publication is part of the project ALIGN4Energy (with project number NWA.1389.20.251) of the research program NWA ORC 2020 which is (partly) financed by the Dutch Research Council (NWO).

* Corresponding author.
*E-mail address:* P.P.VergaraBarrios@tudelft.nl (P.P. Vergara).

## 1. Introduction

Residential Load Profiles (RLPs)[1] have wide applications in areas such as energy supply and demand management [1], modern distribution system planning [2], and risk analysis [3]. The validity of these studies depends largely on the quality of RLPs used. However, access to RLP data is limited due to privacy [4]. RLP generation can provide effective solutions to these problems. On the one hand, distribution system operators (DSOs) rely on RLPs to refine planning decisions. Historical data inaccessibility or limitations can hinder this process, yet the generation of RLPs provides system planners with an alternative to executing more informed planning [5]. For instance, in [6], generated RLPs were used to understand consumption patterns and optimize the system planning. On the other hand, generated RLPs can function as augmented data to support high-level tasks. For example, generated RLPs or PV profiles are used to support the training of models for load prediction [7], non-intrusive load monitoring algorithms [8], and reinforcement learning [9].

Traditional RLP modeling primarily employs Gaussian Mixture Models (GMMs) [10,11]. However, GMMs exhibit limited effectiveness in capturing the complexities of RLP distributions. An alternative approach involves using Copulas models. The study by [12] applies prominent multivariate Copulas models to simulate electric vehicle (EV) charging consumption profiles. Their findings indicate that t-Copulas outperform other functions in modeling these profiles. In recent years, the advancement of Machine Learning (ML) offers new approaches to RLP modeling. In [7,13,14], Generative adversarial networks (GANs) are used to either generate RLPs or PV profiles. In [15], Variational Auto-Encoder (VAE) is proposed to model EV charging profiles. In [16], a hybrid VAE-GAN model is proposed for synthesizing electrical load and PV generation data. The study demonstrates that energy management systems trained on data generated by this model are 8.7% more profitable than the baseline. Similarly, experiments in [17] demonstrate that incorporating generated profiles into battery control algorithm training improves the model's performance bound from approximately 70% to 85%. In [18], a GAN-based generative model is introduced. This model focuses on privacy instead of generating accurate profiles, transforming real-world datasets into high-quality synthetic datasets that ensure user-level privacy. A study by [19] compared the performance of a convolutional Non-linear Independent Component Estimation (NICE) model with GANs in RLP generation. The findings indicate that the convolutional NICE model produces RLPs that exhibit smaller KL divergence relative to real data, suggesting a closer approximation to the actual RLP distributions. In [20], a MultiLoad-GAN was proposed, instead of generating individual RLP, MultiLoad-GAN generates a group of synthetic RLPs which better capture the spatial–temporal correlations among a group of loads. In [21], a diffusion model is proposed, incorporating a folding operation and a novel marginal calibration technique, making it well-suited for high-resolution RLP generation. Despite advancements in RLP generation methods, there remains a lack of comprehensive evaluation approaches for assessing the quality of generated RLPs. To address this gap, [22] recommends a set of fidelity and utility metrics specifically designed for evaluating the quality of smart meter data.

Even though the methods mentioned above show promising results, they do not include the effects of external factors on generated RLPs (such as weather information), which is becoming more important for state-of-the-art generation methods. Conditional generation is a solution to increase the manipulability of models. In [23], conditional Wasserstein GAN (cWGANs) is used for probabilistic load prediction conditions on weather and historical load, which outperforms classical methods such as quantile regression. In [24], a ProfileSR-GAN is proposed for upsampling low-resolution RLPs to high-resolution RLPs. Additionally, incorporating weather information into the generation process was observed to improve results by reducing the Mean Square Error (MSE) of the generated profiles. Specifically, the MSE reduction ranges from 1.3% to 5.6% in experiments. In [25], a conditional VAE (cVAE) is proposed for generating representative load scenarios conditioned on time steps (such as 10:00 am). The study found that the proposed model outperforms traditional methods such as Copulas. In [26], two types of cVAE are employed for generating synthetic energy data conditioned on weather information, the experiments demonstrate that with augmented data, the performance of short-term building energy predictions improved by 12% to 18%. In [27], a transferable flow-based generative model is proposed, which leverages RLP data of different households to improve the prediction of target households. In [28], a conditional multivariate t-distribution (MVT) copula is proposed that outperforms conditional GMMs. In [29], a generative moment matching network (GMMN) is proposed for scenario generation of cooling, heating, and power loads. Results demonstrate that GMMN effectively captures the probability distribution, key load features (such as peaks, ramps, and fluctuations), frequency-domain characteristics, and spatiotemporal correlations. In [30], GMMN is further applied to wind power scenario forecasting. Compared to popular baselines, the generated scenarios better balance sharpness, reliability, and accuracy, demonstrating that WindGMMN is well-suited for wind power forecasting. In [31], a deep generative scenario prediction method based on a redesigned PixelCNN is proposed to model power load uncertainty, demonstrating superior performance in capturing shape, temporal dependencies, and probability distributions compared to VAE, GAN, and NICE. In [32], a deep generative network based on implicit maximum likelihood estimation (IMLE) is proposed for stochastic scenario generation of renewable energy sources and loads. By introducing TransConv layers into the IMLE generator and adopting the Adam optimizer, the method achieves fast and stable convergence, outperforming traditional models such as GANs, VAEs, and Copulas in capturing complex patterns, probability distributions, frequency-domain features, and spatiotemporal correlations. Table 1 summarizes the studies reviewed in this paper.

Despite these developments, current popular conditional generation methods face several challenges. (1) GANs-related models are effective with discrete conditions like days and seasons but perform poorly with continuous variables such as daily or annual consumption, temperature, and irradiation [33,34]. (2) Models such as GANs and VAEs struggle to replicate overall statistical features because they do not directly model probability densities [35]. (3) Copulas models handle continuous conditions well [28], but their lack of scalability makes them impractical for large datasets or high-dimensional data. 4) Flow-based models avoid the above-mentioned limitations but suffer from inadequate modeling capabilities and slow convergence rates [36].

In this paper, we propose a new flow-based generative model architecture coined Full Convolutional Profile Flow (FCPFlow),[2] designed to address challenges previously discussed. The proposed FCPFlow architecture is built upon the idea of a classical flow-based model, proposed initially in [37], but designed to learn the features of RLP data efficiently. The key contributions of this paper are as follows:

- The proposed FCPFlow architecture is designed for RLP generation. Through empirical and theoretical evaluation, FCPFlow demonstrates as main advantages: 1) Enhanced scalability over

[1] In this study, the term *Residential Load Profile* specifically refers to the *net* energy consumption profile of buildings over time, which may or may not include local generation, e.g., photovoltaic generation.

[2] The code, data, and additional materials related to this paper can be found at the following repositories:
(1) Full-Convolutional-Profile-Flow Repository (Personal).
(2) Full-Convolutional-Profile-Flow Repository (TU Delft).

**Table 1**
Summary of literature review.

| Paper | Year | Task type | Model | Target |
|---|---|---|---|---|
| *Unconditional load profile modeling* | | | | |
| [10] | 2009 | Load distribution modeling | GMM | Substation |
| [11] | 2017 | Load profile generation | (Gaussian, t, Gumbel, Clayton, and Frank) Copulas, GMM | Residential building |
| [12] | 2022 | EV load profile generation | (Gaussian, t, Gumbel, Clayton, Clayton, and Frank) Copulas | EV charging station |
| [13] | 2019 | Load profile generation | GAN | Residential building |
| [14] | 2019 | Load profile generation | Recurrent GAN, WGAN, Metropolis–Hastings GAN | Non-residential building |
| [7] | 2020 | Load profile generation | Bidirectional GAN | Commercial and residential building |
| [15] | 2019 | EV load profile generation | VAE | EV charging station |
| [16] | 2023 | Load profile generation | VAE-GAN | Residential building |
| [17] | 2024 | Load profile generation | Gaussian Copulas | Substation |
| [18] | 2022 | Load profile generation | Differentially Private WGAN (DPWGAN) | Residential building |
| [19] | 2023 | Load profile generation | NICE, GAN | Commercial and residential building |
| [20] | 2023 | Load profile generation | Multi-load GAN | Residential building |
| [35] | 2023 | Load profile generation | GMM, GAN, WGAN, WGAN-GP, VAE, Copulas | Residential building and substation |
| [21] | 2024 | Load profile generation | GMM, t-Copula, Diffusion model | Residential building and substation |
| *Conditional load profile modeling* | | | | |
| [23] | 2020 | Probabilistic load prediction | WGAN, CWGAN-GP | Substation |
| [24] | 2022 | Load profile generation | Load profile super-resolution GAN (ProfileSR-GAN) | Residential building |
| [25] | 2022 | Load states generation | cVAE | Country level load |
| [26] | 2022 | Load profile generation | cVAE | Residential building |
| [27] | 2023 | Probabilistic scenario generation | Flow-based model | Residential building |
| [28] | 2021 | Load profile generation | GMM, t-Copulas | Residential building |
| [38] | 2019 | Probabilistic scenario generation | Flow-based model | Residential building |
| [39] | 2022 | Probabilistic scenario generation | WGAN | EV charging station |
| [29] | 2022 | Probabilistic scenario generation | GMMN | Cooling, heating, and power loads |
| [30] | 2021 | Probabilistic scenario generation | GMMN | Wind power |
| [31] | 2022 | Probabilistic scenario generation | PixelCNN | Power load |
| [32] | 2022 | Probabilistic scenario generation | IMLE | Renewable energy sources |

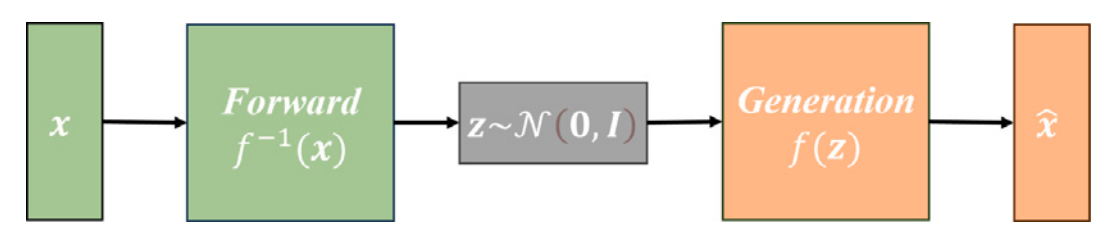

(a) Structure of unconditional flow-based model

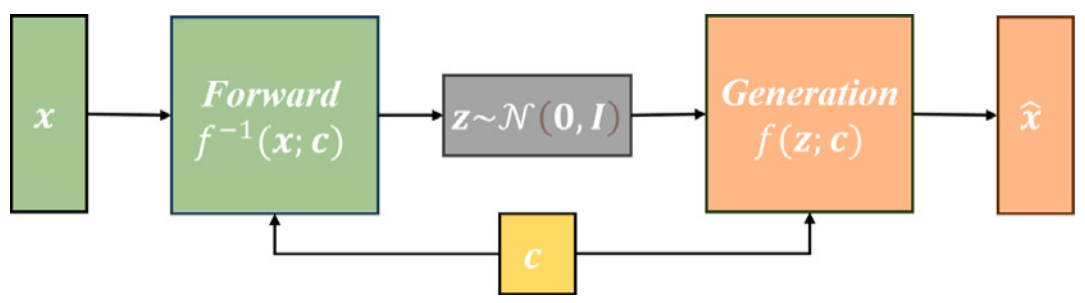

(b) Structure of conditional flow-based model

**Fig. 1.** Structure of flow-based models, where $x$ and $\hat{x}$ are the input data and generated data, $c$ is the condition corresponding to data, $z$ is a latent variable following a standard Gaussian distribution $\mathcal{N}(\mathbf{0}, \boldsymbol{I})$, and $f$ is a bijective function $f : Z \to X$ which is usually constructed by NNs.

traditional statistical models (e.g., GMMs and Copulas), which depend on in-advance defined hypotheses, offering more stable performance across various datasets. 2) Superior modeling performance based on selected evaluation metrics relative to other deep generative models, such as VAE and WGAN with Gradient Penalty (WGAN-GP).

- The proposed FCPFlow architecture is well-suited for RLP generation under continuous conditions (e.g., daily and annual consumption, weather information), which prior research has insufficiently addressed.

## 2. Modeling of residential load profiles

In RLP modeling, a typical daily profile is split into $T$ discrete time steps. For example, an RLP with a resolution of 15 min is characterized by a $T = 96$ time step (24 h), while an RLP with a resolution of 30 min is characterized by a $T = 48$ time step. Each time step corresponds to a specific value of active power consumption in these profiles. In general, a RLP dataset can be described as

$$\mathcal{D} = \{\boldsymbol{x}_i\}_{i=1}^{N} = \{(x_{1,i}, \ldots, x_{T,i})\}_{i=1}^{N}, \tag{1}$$

where $x_{t,i}$ is the active power consumption of $t$th time step, $\boldsymbol{x}_i = (x_{1,i}, \ldots, x_{T,i})$ represents $i$th RLP in $\mathcal{D}$, $N$ is the amount of RLPs in $\mathcal{D}$.

An unconditional deep generative model (e.g., GAN) can be trained to generate RLPs. Such a generative model can be expressed as

$$G_\theta(\boldsymbol{z}_i) = \boldsymbol{x}_i, \tag{2}$$

where $G(\cdot)$ is the generative model which maps $\boldsymbol{z}$ to $\boldsymbol{x}$, $\theta$ is the learnable parameters, $\boldsymbol{z} \sim \pi(\boldsymbol{z})$, and $\pi(\boldsymbol{z})$ can be any simple distribution such as standard Gaussian distribution $\mathcal{N}(\mathbf{0}, \boldsymbol{I})$. Then, a conditional deep generative model can be expressed as

$$G_\theta(\boldsymbol{z}_i; \boldsymbol{c}_i) = \boldsymbol{x}_i, \tag{3}$$

where $\boldsymbol{c}_i = (c_{1,i}, \ldots, c_{B,i})$ is the condition vector corresponding to the $i$th RLP $\boldsymbol{x}_i$. In generative models, conditions can be imposed on the output of the generative model to influence the output outcome. For example, in Section 6.3, $\boldsymbol{c}_i$ represents weather information; therefore, the ML model will generate weather-related RLPs.

## 3. Background

### 3.1. Flow based models

The basic structure of conditional flow-based models is shown in Fig. 1, where $f$ (usually constructed by neural networks (NNs)) is essentially the generator $G_\theta(\cdot)$ in (2) and (3) [36]. During training, function $f^{-1}$ is learned to transform input data $\boldsymbol{x}$ (with condition $\boldsymbol{c}$) into $\boldsymbol{z}$ which follows $\mathcal{N}(\mathbf{0}, \boldsymbol{I})$. Since the function $f^{-1}$ is invertible, once $f^{-1}$ is trained, its inverse $f$ is used to take random samples $\boldsymbol{z}$ (with condition $\boldsymbol{c}$) and generate $\boldsymbol{x}$. In flow-based models, function $f$ (or $f^{-1}$) is usually constructed by stacking multiple invertible transformations $f_i$, meaning $f = f_1 \circ f_2 \ldots \circ f_K$ and $f^{-1} = f_K^{-1} \circ f_{K-1}^{-1} \ldots \circ f_1^{-1}$. Fig. 2 demonstrates how this stacked function transforms a simple $\mathcal{N}(\mathbf{0}, \boldsymbol{I})$ into data distribution $p_K(\boldsymbol{x}|\boldsymbol{c})$ and vice versa. By stacking invertible transformations $f_i$, the modeling capability of $f$ is also increased, enabling function $f$ to simulate more complex data distribution.

**Fig. 2.** Structure of a conditional flow-based model $f = f_1 \circ f_2 \ldots \circ f_K$ which transform standard Gaussian distribution $p_0(z_0) \sim \mathcal{N}(\mathbf{0}, \boldsymbol{I})$ into complex target distribution $p_K(\boldsymbol{x}|\boldsymbol{c})$ [40] and vice versa.

During the training process of a flow-based model, the model parameters can be obtained by maximizing the log-likelihood of the latent variable $\boldsymbol{z}$ with respect to distribution $\mathcal{N}(\mathbf{0}, \boldsymbol{I})$. To do this, the *Change of Variable Theorem* is used, which is expressed as [41]

$$p_X(\boldsymbol{x}|\boldsymbol{c}) = p_Z(\boldsymbol{z}) \left| \det \left( \frac{\partial f^{-1}(\boldsymbol{x};\boldsymbol{c})}{\partial \boldsymbol{x}} \right) \right|, \tag{4}$$

where $\boldsymbol{x}$ is the input RLP data, $\boldsymbol{c}$ is the condition (e.g., weather information), $\boldsymbol{z}$ is a latent variable that follows $\mathcal{N}(\mathbf{0}, \boldsymbol{I})$, $p_X$ and $p_Z$ are the distributions of $\boldsymbol{x}$ and $\boldsymbol{z}$, respectively, and $f$ is a bijective function $f : Z \rightarrow X$ which is usually constructed by NNs, $\det(\cdot)$ is the determinant function. The *Change of Variable Theorem* defines the relation between two distributions if there exists a bijective mapping $f : Z \rightarrow X$. Based on (4), the log-likelihood of $p_X(\boldsymbol{x}|\boldsymbol{c})$ can be expressed as

$$\log p_X(\boldsymbol{x}|\boldsymbol{c}) = \log p_Z(\boldsymbol{z}) + \log \left| \det \left( \frac{\partial f^{-1}(\boldsymbol{x};\boldsymbol{c})}{\partial \boldsymbol{x}} \right) \right|. \tag{5}$$

However, as $f$ is usually constructed by multiple transformations, i.e., $\boldsymbol{x} = f(\boldsymbol{z};\boldsymbol{c}) = f_1 \circ f_2 \circ \ldots \circ f_K(\boldsymbol{z};\boldsymbol{c})$, expression (5) can be further written as

$$\log p_X(\boldsymbol{x}|\boldsymbol{c}) = \log p_Z(z_0) + \sum_{j=1}^{K} \log \left| \det \left( \frac{\partial f_j^{-1}(z_j;\boldsymbol{c})}{\partial z_j} \right) \right|, \tag{6}$$

where $z_j = f_j(z_{j-1};\boldsymbol{c})$ for $j = 1, \ldots, K$ with $z_K = \boldsymbol{x}$, and $z_i$ represents the intermediate latent variable at the $i$th step of the transformation. Thus, the optimal model parameters $\hat{\theta}$ can be obtained by maximizing the log-likelihood of $p_X(\boldsymbol{x}|\boldsymbol{c})$, as

$$\hat{\theta} = \arg\max_{\theta} \; \log p_Z(z_0) + \sum_{j=1}^{K} \log \left| \det \left( \frac{\partial f_j^{-1}(z_j;\boldsymbol{c})}{\partial z_j} \right) \right|. \tag{7}$$

In (6) and (7), $f(\cdot)$, possibly constructed as $f_1 \circ f_2 \circ \ldots \circ f_K(\cdot)$, is essentially the generator $G_\theta(\cdot)$ in (3), where $\theta$ represents the learnable parameters in $f$, such as parameters in NNs.

### 3.2. Combining coupling layer

To guarantee that the transformation $f_i$ is invertible when implemented through NNs, combining coupling layers [42] can be used, denoted as $f_{ccl}$. Fig. 3 shows the structure of $f_{ccl}$ and its inverse $f_{ccl}^{-1}$. The forward process of combining coupling layers $f_{ccl}^{-1}$ can be expressed as

$$\boldsymbol{x}_o, \boldsymbol{x}_e = Split(\boldsymbol{x}) \tag{8}$$

$$(\boldsymbol{x}_e, \boldsymbol{c}) = Concat(\boldsymbol{x}_e, \boldsymbol{c}) \tag{9}$$

$$\hat{\boldsymbol{z}}_e = exp(s_1(\boldsymbol{x}_e;\boldsymbol{c})) \odot \boldsymbol{x}_o + t_1(\boldsymbol{x}_e;\boldsymbol{c}) \tag{10}$$

$$(\hat{\boldsymbol{z}}_e, \boldsymbol{c}) = Concat(\hat{\boldsymbol{z}}_{e,c}) \tag{11}$$

$$\hat{\boldsymbol{z}}_o = exp(s_2(\hat{\boldsymbol{z}}_e;\boldsymbol{c})) \odot \boldsymbol{x}_e + t_2(\hat{\boldsymbol{z}}_e;\boldsymbol{c}) \tag{12}$$

$$\hat{\boldsymbol{z}} = Merge(\hat{\boldsymbol{z}}_e, \hat{\boldsymbol{z}}_o), \tag{13}$$

where $\boldsymbol{x}$ corresponds to one RLP, the operation $Split(\cdot)$ partitions the input vector $\boldsymbol{x}$ (or $\boldsymbol{z}$) into two sub-vectors, $\boldsymbol{x}_e$ and $\boldsymbol{x}_o$, corresponding to the even and odd elements of $\boldsymbol{x}$. Functions $s$ and $t$ are NNs, $Concat(\cdot)$ refers to the method used to concatenate the condition $\boldsymbol{c}$ with $\boldsymbol{x}$ (or $\boldsymbol{z}$) as shown in Fig. 3, and $Merge(\cdot)$ is the inverse operation of $Split(\cdot)$, which merges the sub-vectors. The symbol $\odot$ indicates element-wise multiplication. The generation process of combining coupling layers $f_{ccl}$ can be expressed as

$$\boldsymbol{z}_o, \boldsymbol{z}_e = Split(\boldsymbol{z}) \tag{14}$$

$$(\boldsymbol{z}_e, \boldsymbol{c}) = Concat(\boldsymbol{z}_e, \boldsymbol{c}) \tag{15}$$

$$\hat{\boldsymbol{x}}_e = (\boldsymbol{z}_o - t_2(\boldsymbol{z}_e;\boldsymbol{c}))/exp(s_2(\boldsymbol{z}_e;\boldsymbol{c})) \tag{16}$$

$$(\hat{\boldsymbol{x}}_e, \boldsymbol{c}) = Concat(\hat{\boldsymbol{x}}_e;\boldsymbol{c}) \tag{17}$$

$$\hat{\boldsymbol{x}}_o = (\boldsymbol{z}_e - t_1(\hat{\boldsymbol{x}}_e;\boldsymbol{c}))/exp(s_1(\hat{\boldsymbol{x}}_e;\boldsymbol{c})) \tag{18}$$

$$\hat{\boldsymbol{x}} = Merge(\hat{\boldsymbol{x}}_e, \hat{\boldsymbol{x}}_o). \tag{19}$$

As previously discussed, to obtain the optimal set of parameters of NNs $s$ and $t$, Eq. (6) or (7) can be used. To do this, the log-determinant of $\frac{\partial f_{ccl}^{-1}(\boldsymbol{x};\boldsymbol{c})}{\partial \boldsymbol{x}^{\mathrm{T}}}$ is needed, which can be expressed as

$$\log|\det\left(\frac{\partial f_{ccl}^{-1}(\boldsymbol{x};\boldsymbol{c})}{\partial \boldsymbol{x}^{\mathrm{T}}}\right)| = \log|\det\begin{pmatrix} \boldsymbol{I} & \mathbf{0} \\ * & \exp(s_2(\boldsymbol{z}_e, c)) \end{pmatrix}| + \log|\det\begin{pmatrix} \boldsymbol{I} & \mathbf{0} \\ * & \exp(s_1(\boldsymbol{x}_e, c)) \end{pmatrix}|, \tag{20}$$

where $\boldsymbol{I}$ is the identity matrix, the symbol $*$ denotes the elements in the lower-left quadrant. These elements are represented by $*$ since they do not influence the value of the log-determinant being considered (as multiplied with $\mathbf{0}$).

Theoretically, by stacking combining coupling layer $f = f_{ccl_1} \circ f_{ccl_2} \circ \ldots \circ f_{ccl_K}$, we can build a model that can simulate complex RLPs distribution $p(\boldsymbol{x}|\boldsymbol{c})$, thus enabling generation of conditioned RLPs. However, this modeling approach does not produce satisfactory results for time-series data. The reason is that the classical flow-based requires stacking many layers (with many parameters) to have sufficient modeling capabilities to be able to learn the complex distribution $p(\boldsymbol{x}|\boldsymbol{c})$. This brings two problems: (1) a significantly slow training process and (2) a need for large datasets to support the training process of large models. The proposed FCPFlow model addresses this issue when modeling RLP data while retaining the advantages of flow-based models.

## 4. Proposed model: Full convolutional profile flow

The proposed architecture is composed of multiple FCPFlow blocks, denoted as $f_{fcp_i}$. At the same time, each $f_{fcp_i}$ block is composed of three distinct components: an invertible normalization layer $f_{norm_i}$, an invertible linear layer $f_{lin_i}$, and a combining coupling layer $f_{ccl_i}$, as shown in Fig. 4. The introduction of $f_{lin_i}$ and $f_{norm_i}$ marks the difference from a traditional flow-based model, enriching the modeling capabilities of the FCPFlow architecture to handle time series data such as RLPs. Therefore, the operation of each transformation can be mathematically represented as a composition of these three layers:

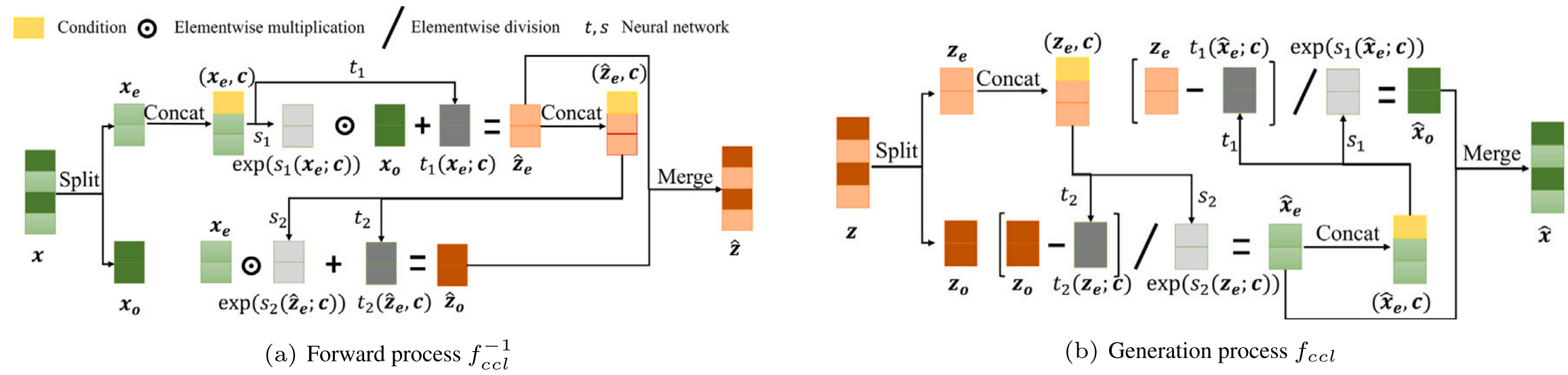

(a) Forward process $f_{ccl}^{-1}$

(b) Generation process $f_{ccl}$

**Fig. 3.** The conditional transformation architecture of combining coupling layer $f_{ccl}$ and its inverse. $t$ and $s$ are the neural networks that can be expressed as $t(\cdot)$ and $s(\cdot)$, both $s$ and $t$ reduce the input dimensions as demonstrated in the figure. $\boldsymbol{x}$ and $\hat{\boldsymbol{x}}$ are the sampled and generated data, respectively. $\boldsymbol{z}$ and $\hat{\boldsymbol{z}}$ are the sampled and generated latent variables, respectively.

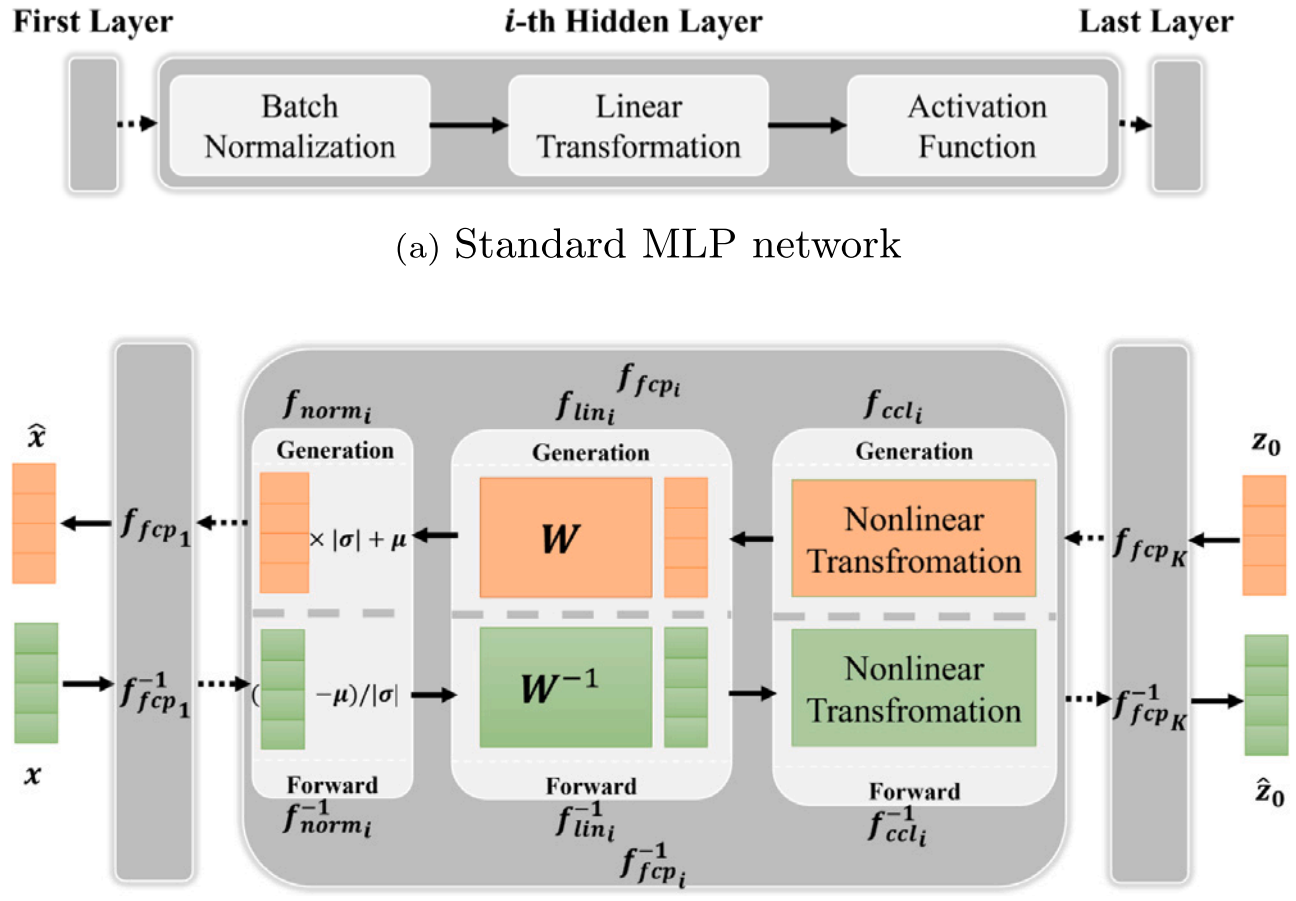

(a) Standard MLP network

(b) FCPFlow network

**Fig. 4.** FCPFlow architecture is structured to process data through a series of FCPFlow blocks. $\boldsymbol{x}$ and $\hat{\boldsymbol{x}}$ are the sampled and generated RLPs, respectively. $\boldsymbol{z}_0$ and $\hat{\boldsymbol{z}}_0$ are the sampled and generated latent variables, respectively. $\sigma$ and $\beta$ are parameters in $f_{norm_i}$ introduced in Section 4.1, $W$ is a matrix that linearly transforms the input vector into another vector, and $W$ is introduced in Section 4.3.

$f_{fcp_i} = f_{norm_i} \circ f_{lin_i} \circ f_{ccl_i}$. One FCPFlow model with $K$ transformations can be expressed as $F^{fcp} = f_{fcp_1} \circ f_{fcp_2} \circ \ldots \circ f_{fcp_K}$. Considering this, each FCPFlow block can be perfectly understood as an invertible counterpart to the traditional multilayer perception (MLP) as depicted in Fig. 4.

For each FCPFlow block, $f_{ccl_i}$, introduced in Section 3, can be understood as an invertible nonlinear transformation (with learnable parameters) that has the same function as the activation function in MLP. $f_{norm_i}$ can be conceptualized as the invertible counterpart of the traditional batch normalization layer. This layer maintains dynamic mean and variance estimates throughout the training phase. Once the FCPFlow model is trained, these estimated parameters (mean and variance) are used in the forward and generation process. In parallel, $f_{lin_i}$ is introduced as an invertible linear transformation layer characterized by learnable parameters $\boldsymbol{W}$. During the forward operation, $f_{lin_i}^{-1}$ performs a linear mapping by applying matrix multiplication between $\boldsymbol{W}^{-1}$ and the layer's input. During the generation operation, $f_{lin_i}$ mirrors this operation by using matrix $\boldsymbol{W}$, thus maintaining the invertibility of the model and facilitating the generation process.

By introducing $f_{norm_i}$ and $f_{lin_i}$, the proposed FCPFlow has higher modeling capabilities than classic flow-based RLP models. To achieve such capabilities, $f_{norm_i}$ aims to stabilize the training process, while $f_{lin_i}$ uses matrix $\boldsymbol{W}$ aiming to understand the correlations among individual time steps of the RLP data. By working with the nonlinear transformations provided by $f_{ccl}$, the proposed FCPFlow blocks can accurately describe the complex, high-dimensional correlations inherent in time series RLP data, addressing the classical flow-based models' limitations mentioned in Section 3.2. To finalize the description of the proposed FCPFlow, the log-determinants of each layer are required. These are introduced in the following sections. Additionally, we provide a simple computation example in Appendix A.1 to help readers better understand the forward computation process of the FCPFlow. The backward process can then be naturally obtained as the inverse of the forward process.

### 4.1. Invertible normalization layer

The functionality of the invertible normalization layer $f_{norm_i}$ is presented in Fig. 4. This normalization operation can be mathematically expressed as in (21) and (22) for the forward and generation processes, respectively.

$$f_{norm}^{-1} \quad \boldsymbol{z} = \frac{\boldsymbol{x} - \boldsymbol{\mu}}{\sqrt{\boldsymbol{\sigma}^2 + \epsilon}} \tag{21}$$

$$f_{norm} \quad \boldsymbol{x} = \boldsymbol{z} \cdot \sqrt{\boldsymbol{\sigma}^2 + \epsilon} + \boldsymbol{\mu}, \tag{22}$$

where $\boldsymbol{\mu}$ and $\boldsymbol{\sigma}$ are mean and standard deviation of $\boldsymbol{x}$ and have the same shape as $\boldsymbol{x}$ and $\boldsymbol{z}$, while $\epsilon$ is a small constant ensuring numerical stability.

To compute the log-likelihood, as expressed in (6), the log-determinant of the invertible normalization layer $f_{norm}^{-1}$ is required, which can be expressed as

$$\log|\det\left(\frac{\partial f_{norm}^{-1}(\boldsymbol{x})}{\partial \boldsymbol{x}^{\mathrm{T}}}\right)| = -\log(|\prod_{i=1}^{T}(|\sigma_i| + \epsilon)|), \tag{23}$$

where $T$ is the length of vector $\boldsymbol{\sigma}$, $\sigma_i$ denoting the $i$th element of vector $\sigma$, for $i \in 1, 2, \ldots, T$.

### 4.2. Invertible linear layer

The operation of the invertible normalization layer, denoted as $f_{lin_i}$, is also presented in Fig. 4. The mathematical formalism for this layer's functionality, in the context of both forward and generative processes, is presented through (24) and (25), respectively.

$$f_{lin}^{-1} \quad \boldsymbol{z} = \boldsymbol{W}^{-1}\boldsymbol{x}, \tag{24}$$

$$f_{lin} \quad \boldsymbol{x} = \boldsymbol{W}\boldsymbol{z}, \tag{25}$$

where $\boldsymbol{W}$ is a invertible matrix. The log-determinants of the invertible linear layer $f_{lin}$ is expressed as

$$\log|\det(\frac{\mathrm{d}f_{lin}^{-1}(\boldsymbol{x})}{\mathrm{d}\boldsymbol{x}^{T}})| = \log|\det(\boldsymbol{W}^{-1})|. \tag{26}$$

### 4.3. Maximum likelihood estimation of FCPFlow

Using the log-determinants of $f_{lin_i}$, $f_{norm_i}$, and $f_{ccl_i}$ (described previously in Section 3.2), the log-likelihood of a FCPFlow model $F^{fcp}$ of $K$ blocks can be expressed as

$$\log p_X(\boldsymbol{x}|\boldsymbol{c}) = \log p_Z(\boldsymbol{z}_0) + \sum_{j=1}^{K}\left(\log\left|\det\left(\frac{\partial f_{fcp_j}^{-1}(\boldsymbol{z}_j;\boldsymbol{c})}{\partial \boldsymbol{z}_j^{\mathrm{T}}}\right)\right|\right)$$

**Fig. 5.** The training process of FCPFlow involves computing the loss based on Eq. (27). The model is trained in the forward process, which involves mapping $\boldsymbol{x}$ to $\hat{\boldsymbol{z}}$, where $\hat{\boldsymbol{z}}$ is expected to follow the distribution $\mathcal{N}(\mathbf{0}, \boldsymbol{I})$. Once the model is trained, we sample $\boldsymbol{z}$ from $\mathcal{N}(\mathbf{0}, \boldsymbol{I})$ and generate $\hat{\boldsymbol{x}}$ during the generation process as shown in Fig. 4.

**Table 2**
Datasets used for the model comparison.

| Country | Resolution | Amount of RLPs | Amount of households | Date range |
|---|---|---|---|---|
| *Unconditional generation* | | | | |
| GE | 15 min | 2131 | 6 | 12.2014–05.2019 |
| *Conditional generation* | | | | |
| NL | 60 min | 27,757 | 82 | 01.2013–12.2013 |
| AUS [43] | 30 min | 10,000 | 156 | 01.2012–12.2012 |
| UK | 30 min | 10,000 | 261 | 01.2013–12.2013 |
| USA [44] | 15 min | 9110 | 73 | 01.2014–12.2018, 05.2019–10.2019 |
| UK weather [45] | 30 min | 10,000 | 261 | 01.2013–12.2013 |
| *Scenarios generation* | | | | |
| NL | 60 min | 365 | 1 | 01.2013–12.2013 |
| UK | 30 min | 365 | 1 | 01.2013–12.2013 |
| USA [44] | 15 min | 365 | 1 | 01.2018–12.2018 |
| *Computational cost* | | | | |
| NL | 60 min | 364 | 1 | 01.2013–12.2013 |
| *Peak analysis* | | | | |
| Same as conditional generation | | | | |
| *Data requirement analysis* | | | | |
| NL | 60 min | 364 | 1 | 01.2013–12.2013 |

$$= \log p_Z(\boldsymbol{z}_0) + \sum_{j=1}^{K} \left( \log \left| \det \left( \frac{\partial f^{-1}_{ccl_j} \circ f^{-1}_{lin_j} \circ f^{-1}_{norm_j}(\boldsymbol{z}_j; \boldsymbol{c})}{\partial \boldsymbol{z}_j^{\mathrm{T}}} \right) \right| \right). \tag{27}$$

Given that the log-likelihood of $F^{fcp}$ can be calculated using the expression in (27), it becomes feasible to train model $F^{fcp}$ through the application of gradient descent maximizing such log-likelihood. In this case, the parameters subject to optimization include the parameters of matrices $\boldsymbol{W}_i$ within $f_{lin_i}$, alongside the parameters of the NNs, denoted as $s_i$ and $t_i$ from $f_{ccl_i}$. The training process of the model is shown in Fig. 5.

## 5. Simulations setup

### 5.1. Implementation details

A notable challenge in training the proposed PCPFlow arises from using exponential and logarithmic functions in the combining coupling layer $f_{ccl_i}$, which can lead to numerical instability. To mitigate this issue, we implement a soft clamping mechanism in combining coupling layers, as suggested by [42]. The trick is simply replacing $s(\boldsymbol{x}_i; \boldsymbol{c})$ with $s^{clamp}(\boldsymbol{x}_i; \boldsymbol{c})$ in $f_{ccl_i}$ (shown in Fig. 3 as $s_1(\cdot)$ and $s_2(\cdot)$) , which is mathematically expressed as

$$s^{clamp}(\boldsymbol{x}_i; \boldsymbol{c}) = \frac{2\alpha}{\pi} acrtan(\frac{s(\boldsymbol{x}_i; \boldsymbol{c})}{\alpha}), \tag{28}$$

where $\alpha$ is a hyper-parameter, $s^{clamp}(\boldsymbol{x}_i; \boldsymbol{c}) \approx s(\boldsymbol{x}_i; \boldsymbol{c})$ for $|s(\boldsymbol{x}_i; \boldsymbol{c})| \ll \alpha$ and $s^{clamp}(\boldsymbol{x}_i; \boldsymbol{c}) \approx \pm\alpha$ for $\alpha \ll |s(\boldsymbol{x}_i; \boldsymbol{c})|$. $s^{clamp}(\boldsymbol{x}_i; \boldsymbol{c})$ can effectively curb the potential instabilities caused by the exponential function $\exp(s^{clamp}(\boldsymbol{x}_i; \boldsymbol{c}))$ [42]. Based on our experiment, the best range of $\alpha$ is $(0.1, 1)$.

### 5.2. Data introduction

For a comprehensive comparison, RLP datasets from five countries were used. Table 2 outlines the details of these datasets. The UK, NL, and GE datasets sources can be found in our previous work [35]. The NL, UK, AUS, and USA datasets are used for conditional generation, in which the conditions are annual and daily total consumption in kWh. UK weather dataset is also used for the conditional generation, in which the conditions are different weather information (including cloud cover, sunshine, irradiation, maximum temperature, minimum temperature, mean temperature, pressure, and precipitation). The UK, NL, and USA datasets are also used for scenario generation experiments. Moreover, the NL dataset is further applied in deeper analyses, including computational cost analysis and data requirement analysis. The number of RLPs in Table 2 refers to the amount of data used for the experiments. One RLP is defined as the consumption profile of a family for a day (with different time resolutions). For example, a one-week consumption profile for two households equals $7 \times 2 = 14$ RLPs.

### 5.3. Evaluation metrics

#### 5.3.1. Evaluation metrics for evaluating overall (conditional) RLP generation performance

Aligning with [25,28], the evaluation metrics used in this paper from Sections 6.1 to 6.3 are Energy Distance (ED), Maximum Mean Discrepancy (MMD), Wasserstein Distance (WD), KS Distance (KS), and MSE of Autocorrelation (MSE.A). Among these metrics, ED, WD, and KS assess the overall distributional differences between the generated RLPs and the original dataset, while MSE.A measures the differences in linear temporal correlations. MMD, on the other hand, captures high-level statistical features, such as non-linear correlations, that are not reflected by the other metrics. The smaller the value of the above metrics, the better the performance of the model. The ED between two distributions $P$ and $Q$ can be represented as follows

$$D_E(P,Q) = 2\mathbb{E}\|\boldsymbol{x}-\boldsymbol{y}\| - \mathbb{E}\|\boldsymbol{x}-\boldsymbol{x}'\| - \mathbb{E}\|\boldsymbol{y}-\boldsymbol{y}'\| \tag{29}$$

where $\boldsymbol{x}$ and $\boldsymbol{x}'$ represent independent RLPs sampled from real distribution $P$, $\boldsymbol{y}$ and $\boldsymbol{y}'$ represent independent generated RLPs sampled from distribution $Q$ ($Q$ represents the distribution of generated RLPs). The KS between two empirical cumulative distribution functions (CDF) $F(\boldsymbol{x})$ and $F(\boldsymbol{y})$ is given by

$$D_{KS}(F(\boldsymbol{x}), F(\boldsymbol{y})) = \sup_{\boldsymbol{x}} |F(\boldsymbol{x}) - F(\boldsymbol{y})|, \tag{30}$$

where sup denotes the supremum over all possible values of $\boldsymbol{x}$. For real and generated RLP datasets $\mathcal{D}_r$ and $\mathcal{D}_g$, the MSE.A is computed as

$$\text{MSE} = \sum (R(\mathcal{D}_r) - R(\mathcal{D}_g))^2, \tag{31}$$

where $R(\mathcal{D}_r), R(\mathcal{D}_g)$ represent the autocorrelation of two datasets. The WD between two probability measures is defined as

$$W(P,Q) = \inf_{\pi \in \Pi(P,Q)} \int_{\mathbb{R}^d \times \mathbb{R}^d} \|\boldsymbol{x}-\boldsymbol{y}\| \, \mathrm{d}\pi(\boldsymbol{x},\boldsymbol{y}), \tag{32}$$

where $W(P,Q)$ is the WD between two distribution $P$ and $Q$, $\boldsymbol{x}$ and $\boldsymbol{y}$ are RLPs sampled from $P$ and $Q$. The MMD is expressed as

$$\text{MMD}(P,Q) = \sqrt{\mathbb{E}[k(\boldsymbol{x},\boldsymbol{x}')] + \mathbb{E}[k(\boldsymbol{y},\boldsymbol{y}')] - 2\mathbb{E}[k(\boldsymbol{x},\boldsymbol{y})]}, \tag{33}$$

where $k$ is the Gaussian kernel $k(\boldsymbol{x},\boldsymbol{x}') = \exp\left(-\frac{\|\boldsymbol{x}-\boldsymbol{x}'\|^2}{2\sigma^2}\right)$.

### 5.3.2. Evaluation metric for peaks generation

Peak consumption is a critical factor for distribution system planning and operation. To assess the models' performance in accurately capturing peak consumption and corresponding times, we apply the metric described in [35], where each day's peak consumption and its corresponding time are represented as a point of the form *(time, peak)*. The centers of the *(time, peak)* points for both the real RLP data and the RLPs generated by the models are then calculated (over a defined period, e.g., for a year). The proximity between the real data centers and the generated RLP data centers is measured using Euclidean distance, which quantitatively assesses the models' overall performance in time-peak modeling. This metric can be expressed as

$$EuD = \sqrt{(\bar{p}_\mathrm{r} - \bar{p}_\mathrm{g})^2 + (\bar{t}_\mathrm{r} - \bar{t}_\mathrm{g})^2}, \tag{34}$$

where $\bar{t}_\mathrm{r}$ and $\bar{p}_\mathrm{r}$ represent the coordinates (time and peak) of the center of the real RLP data, and $\bar{t}_\mathrm{g}$ and $\bar{p}_\mathrm{g}$ represent the coordinates of the center of the generated RLP data.

The Mean Absolute Percentage Error (MAPE) is also used to assess a model's ability to accurately generate seasonal peak values, critical during distribution networks planning [46]. The MAPE is defined as

$$MAPE_s = 100 \times \left| \frac{\sum_{i=1}^{N} x^s_{i,\mathrm{peak}} - \sum_{i=1}^{N} y^s_{i,\mathrm{peak}}}{\sum_{i=1}^{N} y^s_{i,\mathrm{peak}}} \right|, \tag{35}$$

where $N$ is the number of data points of the $s$th season in the evaluation period, and $x^s_{i,\mathrm{peak}}$ and $y^s_{i,\mathrm{peak}}$ denote the generated and actual peak values for the $s$th season, respectively. In this paper, we evaluate MAPE only for summer and winter, which are more significant for distribution networks planning, $s \in \{\text{summer, winter}\}$.

### 5.3.3. Evaluation metrics for probabilistic scenario generation

The evaluation metrics for the experiments in Section 7 are Pinball loss (PL), Continuous Ranked Probability Score (CRPS), and the MSE between the true and the average of generated scenarios [23,38]. Similarly, the smaller the value of these metrics, the better the generation results of the model. The MSE is simply defined by the MSE of true value $\boldsymbol{y}_t$ and the average of generated scenarios $\overline{\boldsymbol{y}_p}$. The PL function, used in quantile regression, is defined as

$$L_\tau(\boldsymbol{y}_t, \boldsymbol{y}_p) = \begin{cases} \tau(\boldsymbol{y}_t - \boldsymbol{y}_p) & \text{if } \boldsymbol{y}_t > \boldsymbol{y}_p \\ (1-\tau)(\boldsymbol{y}_p - \boldsymbol{y}_t) & \text{otherwise,} \end{cases} \tag{36}$$

where $\boldsymbol{y}_p$ is the generated scenarios, $\tau$ is the quantile (e.g., 0.9 for the 9th percentile).

The CRPS is given by the integral of the squared difference between the CDF of the generated distribution and the observed real value's CDF. The CRPS is expressed as

$$\text{CRPS}(F, \boldsymbol{y}_t) = \int_{-\infty}^{\infty} \left(F(\boldsymbol{y}_p) - \mathbb{1}(\boldsymbol{y}_p \geq \boldsymbol{y}_t)\right)^2 \mathrm{d}\boldsymbol{y}_p, \tag{37}$$

where $\text{CRPS}(F, \boldsymbol{y}_t)$ is the CRPS for a prediction distribution $F$, $F(\boldsymbol{y}_p)$ represents the CDF of the predicted distribution evaluated at $\boldsymbol{y}_p$, and $\mathbb{1}(\boldsymbol{y}_p \geq \boldsymbol{y}_t)$ is the indicator function, which equals 1 when $\boldsymbol{y}_p \geq \boldsymbol{y}_t$ and 0 otherwise.

**Table 3**
Results of evaluation metrics for GE dataset.

| Model | ED | MSE.A | KS | WD | MMD |
|---|---|---|---|---|---|
| t-Copulas | 0.1212 | 0.0134 | 0.2956 | 0.0624 | 0.0190 |
| GMMs | 0.0394 | 0.0224 | 0.0861 | 0.0168 | 0.0201 |
| WGAN-GP | 0.0543 | 0.0124 | 0.1527 | **0.0110** | 0.0237 |
| **VAE** | 0.0499 | 0.0084 | 0.1331 | 0.0196 | 0.0161 |
| **DDPM** | 0.0468 | 0.0098 | **0.0851** | 0.0125 | 0.0178 |
| **FCPFlow** | **0.0372** | **0.0053** | 0.1057 | 0.0147 | **0.0068** |

### 5.4. Hyperparameter setting

We utilize cyclical learning rates [47] for the training process, with the highest and lowest learning rates set to $1e{-}3$ and $1e{-}5$ for all generation experiments. We simply set the learning rate for other experiments to be $1e{-}3$. Other hyperparameters (such as number of blocks, the width of the model, $\alpha$, and batch size) for different experiments are tuned during the experiments, and the detailed settings of models are available in our repository[2] mentioned before. Additionally, in Appendix A.2, we provide a summary of the benchmark models employed across main experiments, along with the corresponding FCPFlow structures and the parameter scales of the deep generative models.

## 6. Simulation results for RLP generation

### 6.1. Unconditional generation

In this section, we first evaluate the performance of the proposed FCPFlow architecture on the unconditional generation task. Based on previous studies [28,39], t-Copulas, GMM, VAE, Denoising Diffusion Probabilistic Models (DDPM) [48], and WGAN-GP are selected as benchmarks for comparison against FCPFlow. The GE dataset is used for comparison. We select the FCPFlow and WGAN-GP with the smallest ED during the training. The loss curve of models during the training is shown in Appendix A.3. Table 3 summarizes the results of evaluation metrics, where we find that FCPFlow outperforms other models in the ED (decrease by 0.0022), MSE.A (decrease by 0.0041), and MMD (decrease by 0.0093) metrics, and a high position in KS and WD. This suggests its superiority in capturing temporal correlations. In contrast, GMM, although effective at modeling the overall distribution (as shown by small ED and KS scores), falls short in modeling the correlation between time steps. In contrast, deep generative models like VAE, DDPM, WGAN-GP, and FCPFlow perform well in this regard. Additionally, DDPM achieves the lowest KS score (0.0851), and outperforms both VAE and WGAN-GP in most of the metrics. VAE, while not the best in any single metric, shows relatively balanced performance across the different evaluation metrics.

Fig. 6 shows the generated results of different models, it reveals that t-Copulas tend to produce RLPs with higher daily consumption, as indicated by a greater number of RLPs with more intense red hues. This pattern, consistently observed across various experiments, may suggest that t-Copulas-generated RLPs exhibit less volatility (an RLP with relatively high consumption at the first time step tends to keep this high consumption pattern in the following time steps) compared to the original data. Interestingly, from Fig. 6, it appears that DDPM successfully captures the volatility patterns of the original data. However, opposite to the t-Copula model, DDPM tends to generate RLPs with lower daily energy consumption, as indicated by a smaller number of RLPs with intense red hues. In contrast, Fig. 6 also shows that FCPFlow-generated RLPs exhibit volatility patterns more closely aligned with those of the original data.

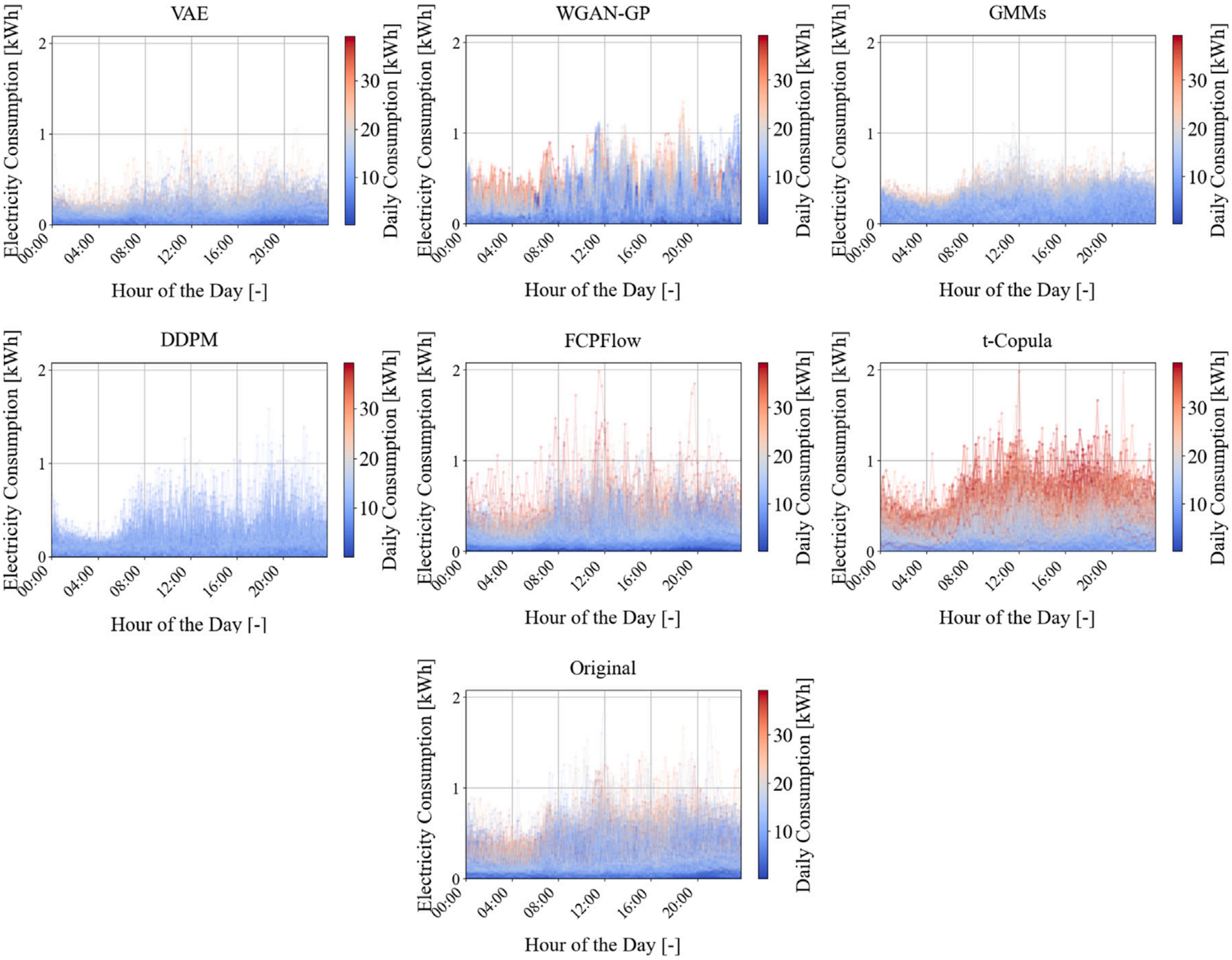

**Fig. 6.** The unconditional generation results of GMM, t-Copulas, WGAN-GP, VAE, DDPM, and FCPFlow on the GE dataset are shown. The color of each RLP represents the total daily consumption, with the color bar on the right indicating the corresponding total daily consumption for each color. The $y$-axis on the left indicates the RLP's electricity consumption at each step.

### 6.2. Conditional generation based on consumption

In this section, we test the FCPFlow's performance on conditional generation. The conditions used are annual consumption $c_{ann}$ and daily consumption $c_{daily}$ (in kWh) related to each RLP. Consequently, the FCPFlow model is formalized as $F(\boldsymbol{z}; c_{ann}, c_{daily})$. The datasets used are UK, AUS, NL, and USA which have different resolutions. Previous research by [28] has established that t-Copulas outperforms conditional GMMs in conditional RLP generation tasks. Therefore, our comparative analysis focuses on measuring the performance differences between FCPFlow , cWGAN-GP, and t-Copulas.

In these experiments, the datasets are split into a test set (20% of the data) and a training set (80% of the data). The models are first trained using the training dataset. Then, RLPs are generated according to the conditions specified in the test set. Finally, we compute the evaluation metrics by comparing these generated RLPs with the actual data in the test set. We select the model with the smallest ED during the training. Fig. 7(a–d) shows the generated results, where we observe again that t-Copulas tend to produce RLPs with higher daily consumption, identifiable by the more vivid red colors in Fig. 7(a) and (c). Interestingly, in Fig. 7(b), the FCPFlow successfully generated some outliers in the test dataset—the RLPs with very high peaks. These outliers are red RLPs with a high consumption peak, t-Copulas fails to generate these outliers. Additionally, cWGAN-GP demonstrates decent generation performance in Fig. 7(a), but its generated profiles in Fig. 7(d) exhibit noticeably reduced volatility.

**Table 4**
Results of evaluation metrics for NL, UK, and USA dataset.

| Model | ED | MSE.A | KS | WD | MMD |
|---|---|---|---|---|---|
| *NL 60 min resolution* | | | | | |
| t-Copulas | 0.1896 | 0.0221 | 0.2831 | 0.1033 | 0.0473 |
| **cWGAN-GP** | 0.0924 | 0.0057 | 0.2319 | 0.0398 | 0.0336 |
| **FCPFlow** | **0.0650** | **0.0051** | **0.1546** | **0.0387** | **0.0150** |
| *UK 30 min resolution* | | | | | |
| t-Copulas | 0.0064 | **0.0003** | 0.0192 | **0.0037** | 0.0048 |
| **cWGAN-GP** | 0.0146 | 0.0014 | 0.0311 | 0.0094 | 0.0027 |
| **FCPFlow** | **0.0052** | 0.0004 | **0.0106** | 0.0038 | **0.0006** |
| *AUS 30 min resolution* | | | | | |
| t-Copulas | **0.0628** | 0.0037 | 0.1290 | **0.0389** | 0.0070 |
| **cWGAN-GP** | 0.0718 | 0.0036 | 0.2387 | 0.0473 | 0.0035 |
| **FCPFlow** | 0.0635 | **0.0013** | **0.1199** | 0.0463 | **0.0010** |
| *USA 15 min resolution* | | | | | |
| t-Copulas | **0.0141** | 0.0019 | **0.0198** | **0.0315** | 0.0016 |
| **cWGAN-GP** | 0.0736 | 0.0032 | 0.0863 | 0.0761 | 0.0021 |
| **FCPFlow** | 0.0320 | **0.0017** | 0.0571 | 0.0601 | **0.0014** |

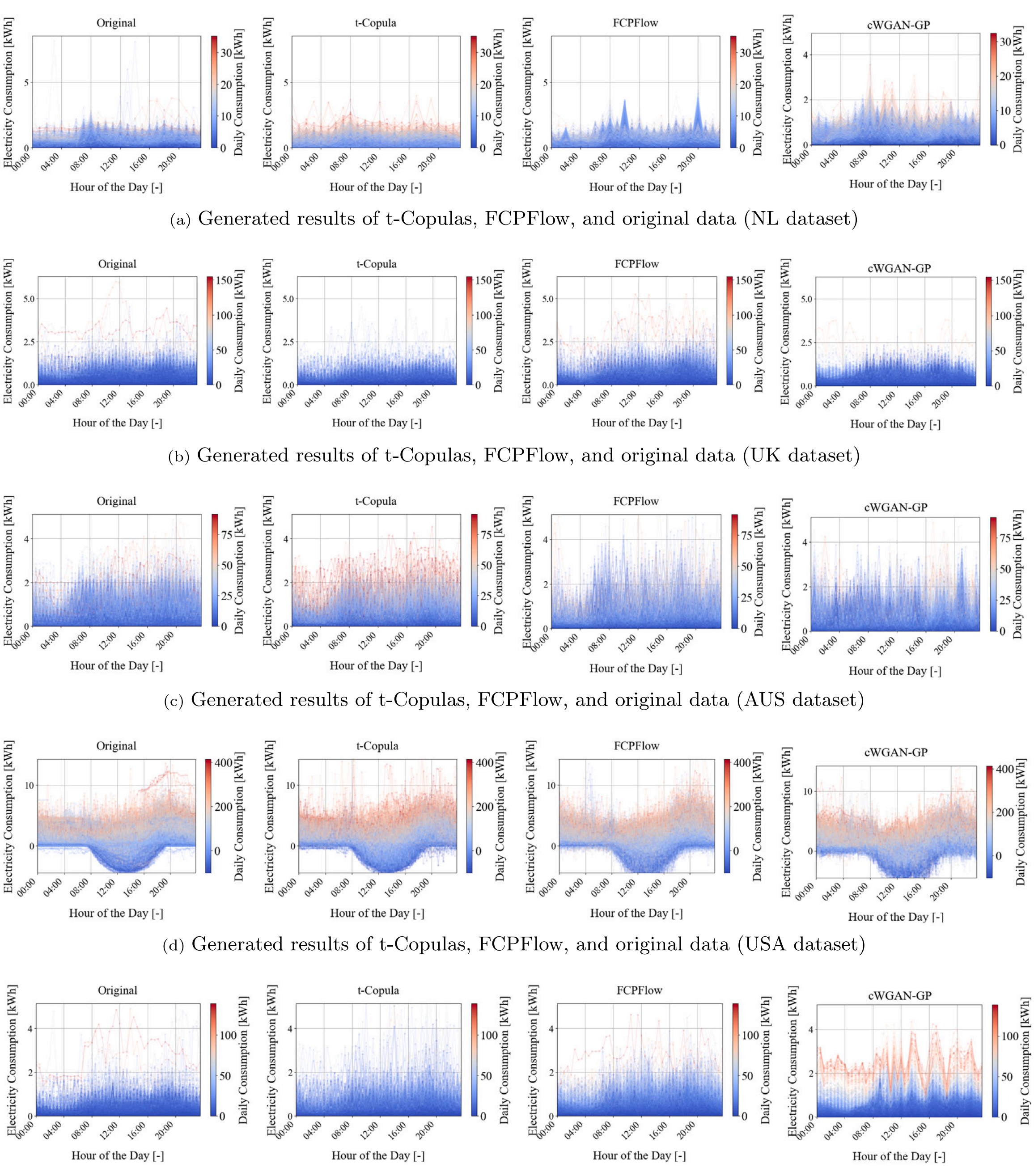

(a) Generated results of t-Copulas, FCPFlow, and original data (NL dataset)

(b) Generated results of t-Copulas, FCPFlow, and original data (UK dataset)

(c) Generated results of t-Copulas, FCPFlow, and original data (AUS dataset)

(d) Generated results of t-Copulas, FCPFlow, and original data (USA dataset)

(e) Generated results of t-Copulas, FCPFlow, and original data (UK weather dataset)

**Fig. 7.** Conditional generated results of t-Copulas, cWGAN-GP, and FCPFlow using UK, USA, and UK-weather datasets. The color of each RLP represents the total daily consumption, with the color bar on the right indicating the corresponding total daily consumption for each color. The *y*-axis on the left shows the RLP's electricity consumption at each step. (For interpretation of the references to color in this figure legend, the reader is referred to the web version of this article.)

In a more quantitative analysis, as shown in Table 4, we observe that, although cWGAN-GP demonstrates reasonable performance on the NL dataset—as indicated by lower ED, MSE.A, KS, WD, and MMD values compared to the t-Copula model, its performance deteriorates significantly as the temporal resolution increases. In particular, it becomes inferior and, in some cases, not comparable to the t-Copula model, especially for the 15-min resolution USA dataset, where cWGAN-GP performs worse across all evaluation metrics. Additionally, FCPFlow outperforms t-Copulas and cWGAN-GP on most metrics across experiments, in which FCPFlow achieves the best MMD in four datasets (100%), and the best MSE.A and KS in three datasets (75%), the best ED in two datasets (50%). FCPFlow's superior performance in modeling correlations can be attributed to its lack of predefined assumptions, whereas t-Copulas relies on the assumption of using the Student-t distribution to model temporal correlations.

Another observation is that the performance of the t-Copulas model is highly dependent on the characteristics of the RLP datasets, a finding that echoes the research presented in [35]. The t-Copulas demonstrates superior performance with the UK dataset, closely matching FCPFlow in several metrics except for the MMD. In the USA dataset, t-Copula achieves a better overall performance. However, t-Copula struggles with the NL and GE datasets, which have a significant performance gap compared with the FCPFlow model. The variation of t-Copula model's performance may be attributed to two reasons: (1) t-Copulas model in [28] relies on the empirical CDF to model the marginal distribution, this method allows t-Copulas to reproduce the marginal distribution of the training set perfectly, but it can also lead to overfitting, which negatively affects the accuracy of the model in representing peaks and correlations in the test set, and (2) t-Copula assumes that the correlation of RLPs follows Student-t distribution, which may be different from reality.

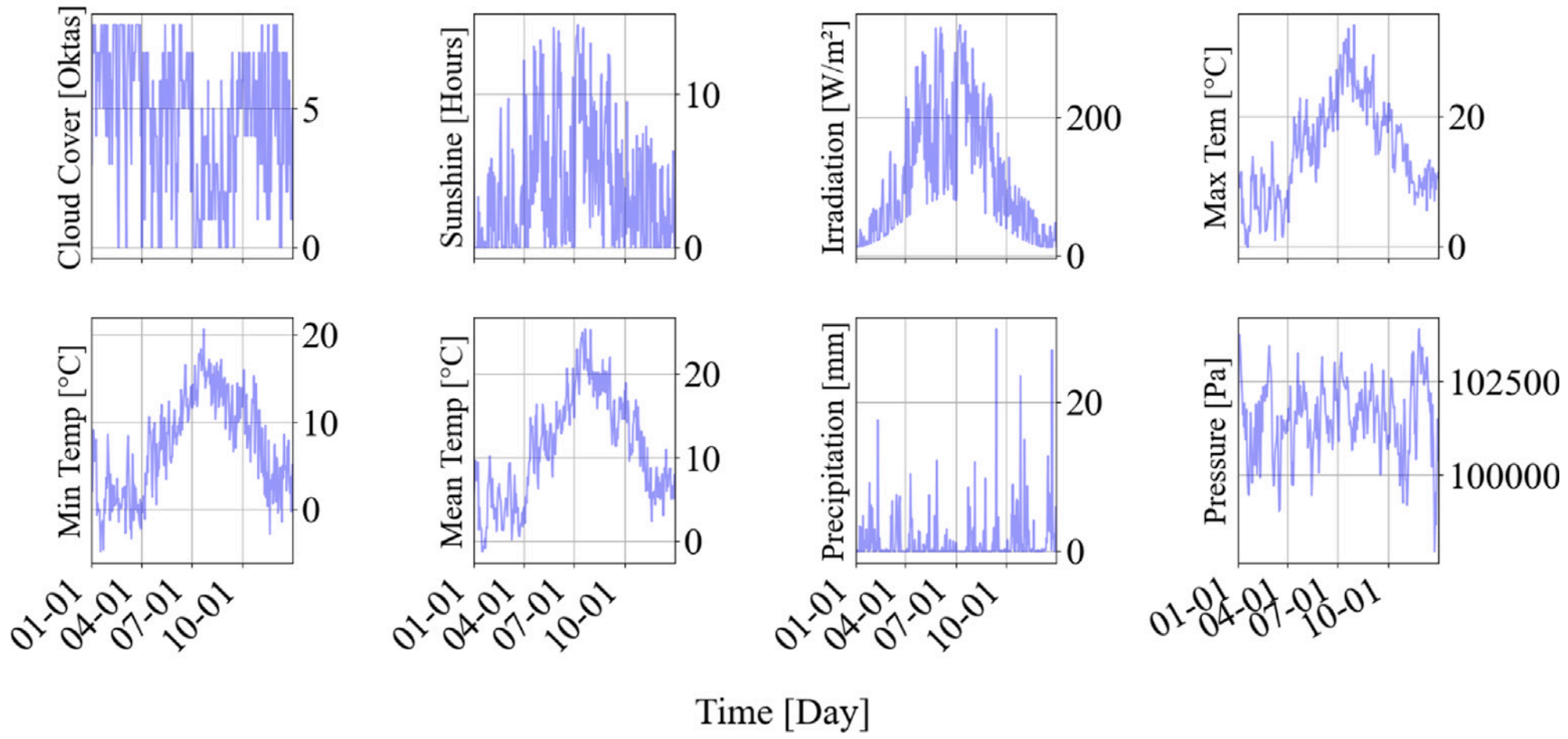

**Fig. 8.** Annual variation of weather conditions, including cloud cover (Oktas), sunshine (Hours), solar irradiation (W/m$^2$), temperature (maximum, minimum, and mean) (°C), pressure (Pa), and precipitation (mm).

**Table 5**
Results of evaluation metrics for UK weather dataset.

| Model | ED | MSE.A | KS | WD | MMD |
|---|---|---|---|---|---|
| *Overall performance* | | | | | |
| t-Copulas | 0.0306 | 0.0037 | **0.0356** | 0.0263 | **0.0024** |
| **cWGAN-GP** | 0.0311 | 0.0157 | 0.0670 | 0.0222 | 0.0117 |
| **FCPFlow** | **0.0267** | **0.0018** | 0.0487 | **0.0202** | 0.0072 |
| *Max temp* > 25 °C | | | | | |
| t-Copulas | 0.0576 | 0.0135 | 0.0753 | 0.0464 | **0.0125** |
| **cWGAN-GP** | 0.0546 | 0.0293 | 0.1187 | 0.0339 | 0.0521 |
| **FCPFlow** | **0.0212** | **0.0053** | **0.0377** | **0.0170** | 0.0131 |
| *Min temp* < 3 °C | | | | | |
| t-Copulas | 0.0348 | 0.0061 | **0.0409** | 0.0295 | **0.0033** |
| **cWGAN-GP** | 0.0511 | 0.0165 | 0.0798 | 0.0396 | 0.0120 |
| **FCPFlow** | **0.0288** | **0.0028** | 0.0561 | **0.0238** | 0.0062 |
| *Irradiation* > 250 W/m$^2$ | | | | | |
| t-Copulas | 0.0432 | 0.0136 | **0.0623** | 0.0357 | **0.0133** |
| **cWGAN-GP** | 0.0321 | 0.0146 | 0.0901 | 0.0181 | 0.0533 |
| **FCPFlow** | **0.0301** | **0.0040** | 0.0664 | **0.0171** | 0.0188 |
| *Sunshine* > 10 h | | | | | |
| t-Copulas | 0.0586 | 0.0127 | 0.0766 | 0.0492 | **0.0109** |
| **cWGAN-GP** | 0.0343 | 0.0139 | 0.0981 | 0.0183 | 0.0514 |
| **FCPFlow** | **0.0252** | **0.0044** | **0.0531** | **0.0148** | 0.0182 |
| *Precipitation*> 10 mm | | | | | |
| t-Copulas | 0.0721 | 0.0275 | 0.0713 | **0.0109** | 0.0399 |
| **cWGAN-GP** | 0.1174 | 0.0608 | 0.1536 | 0.0682 | 0.0533 |
| **FCPFlow** | **0.0170** | **0.0154** | **0.0502** | 0.0628 | **0.0350** |

### 6.3. Conditional generation based on weather

In this section, we examine the performance of FCPFlows using weather information as conditions. The dataset used is UK weather data with a 30-min resolution. As previously done, the data set is divided into a training set (80% of the data) and a test set (the remaining 20%). Weather conditions considered in this analysis include cloud cover (Oktas), sunshine (Hours), solar irradiation (W/m$^2$), temperature (maximum, minimum, and mean) (°C), pressure (Pa), and precipitation (mm). The variation of these conditions throughout the year is shown in Fig. 8. For this experiment, the FCPFlow model is represented as $F(\boldsymbol{z}; c_{ann}, \boldsymbol{c}_{weather})$, where $\boldsymbol{c}_{weather}$ contains eight specified weather features. Similarly, we select the model with the smallest ED during the training.

Fig. 7(e) shows the overall generated results using 100% of the test set. The results show again that FCPFlow generates some outliers with high peaks (the red curve with the high peaks) while t-Copulas and cWGAN-GP fail to capture such extreme cases. Additionally, we observe that cWGAN-GP tends to exhibit a mode collapse phenomenon in Fig. 7(e) and generate less volatile results. A detailed quantitative analysis in Table 5 shows that the FCPFlow model significantly outperforms the t-Copulas and cWGAN-GP model in overall performance. Specifically, enhancements include a reduction of 0.039 in ED, a reduction of 0.019 in MSE.A, and a reduction of 0.061 in WD. Additionally, we evaluated the models' generation performance under relatively extreme weather conditions, defined as maximum temperature >25 °C, minimum temperature <3 °C, irradiation >250 W/m$^2$, sunshine duration >10 hours, and precipitation >10 mm. The objective was to assess whether FCPFlow could maintain its performance under these conditions. The quantitative results are also presented in Table 5. These results once again demonstrate that FCPFlow consistently exhibits lower losses and generally outperforms t-Copulas and cWGAN-GP across most evaluation metrics. In these evaluations, FCPFlow achieves the best ED and MSE.A scores in all six scenarios (100%), the best WD in five out of six scenarios (83%), and the best KS in three out of six cases (50%).

### 6.4. Peak generation analysis

In this section, we conduct a detailed analysis of the models' performance in accurately capturing peak consumption and corresponding times from previous experiments (unconditional generation and conditional generation, as detailed in Sections 6.1, 6.2, and 6.3), using the *(time, peak)* and MAPE metrics as described in Section 5.3.2.

Results in Table 6 demonstrate that FCPFlow outperforms t-Copula in four of five datasets, with better performance in both MAPE (for summer and winter) and Euclidean Distances of the metric *(time, peak)*. However, in the UK weather dataset, FCPFlow exhibits inferior performance. This performance discrepancy can also be observed in Fig. 9(b), where the center of FCPFlow shows more significant deviation from the center of the original data on both the $y$-axis (peak values) and the $x$-axis (corresponding time), compared with t-Copula.

The underperformance of FCPFlow in the UK weather is primarily attributed to deviations in accurately modeling the peak times rather than the peak values, where we can observe in Fig. 9(b) that the difference of centers in the $y$-axis (peak value) is relatively small,

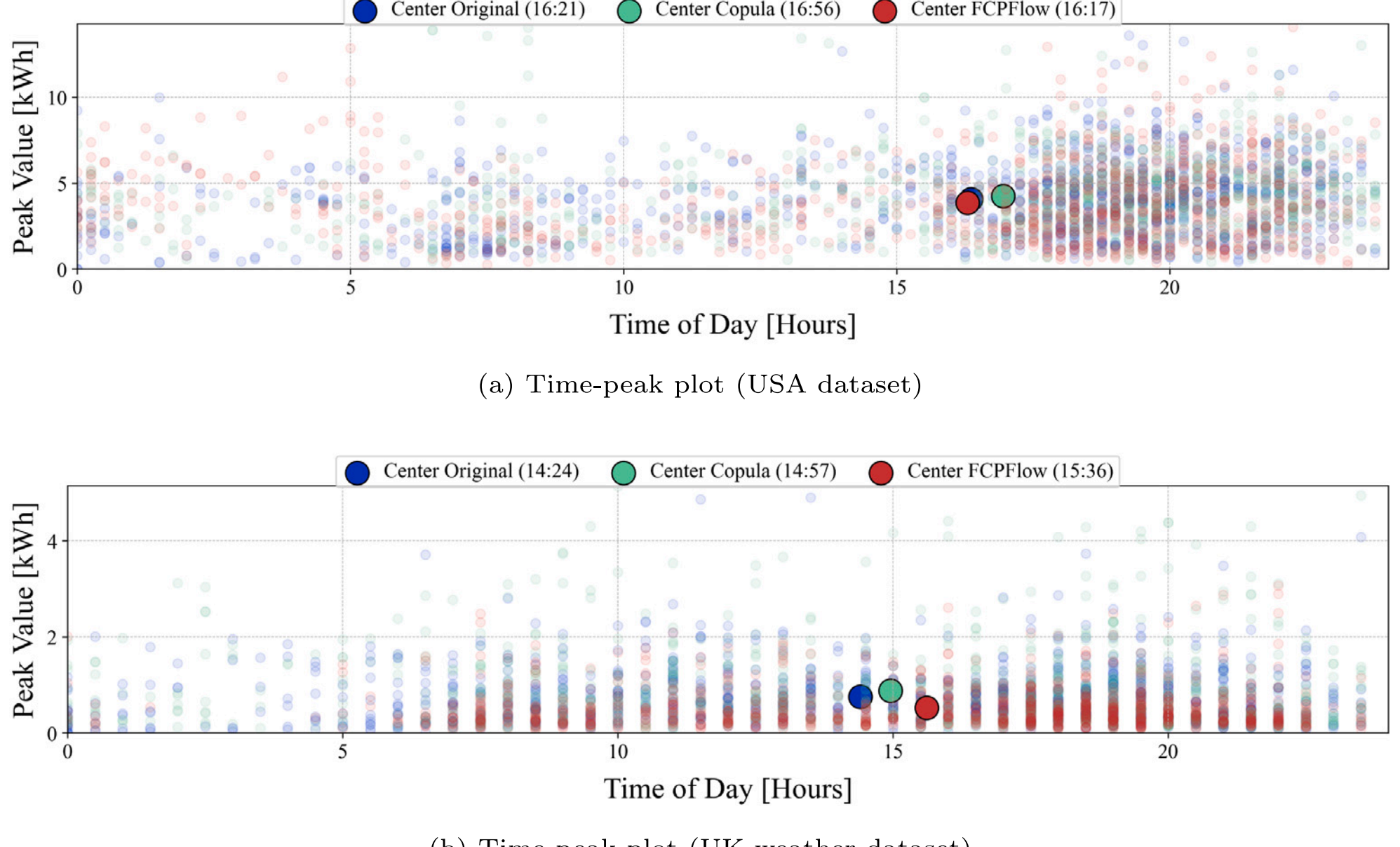

(a) Time-peak plot (USA dataset)

(b) Time-peak plot (UK weather dataset)

**Fig. 9.** The *(time, peak)* plots based on generation experiments illustrate the relationship between daily peak values and their corresponding times. As described in Section 5.3.2, in these figures, the transparent points represent the daily *(time, peak)* observations for one daily load profile. The solid points indicate the centers, calculated as the averages of all *(time, peak)* points of the real datasets and the generated RLP datasets from different models. (For interpretation of the references to color in this figure legend, the reader is referred to the web version of this article.)

while the difference in the $x$-axis (corresponding time) is relatively significant. These deviations, observed in Fig. 9(b), may be linked to a model collapse—a phenomenon in generative modeling where the model fails to fully capture the diversity of the target data distribution, often leading to repetitive or biased outputs [49]. Our previous work [35] analyzes this phenomenon in RLP generation. Model collapse is a common challenge in generative models. In the UK weather dataset, most conditions are likely to produce peaks between 8:00 and 24:00, which is considered a higher probability of occurrence. As a result, the model may overlook generating peaks between 1:00 and 5:00, which have a lower likelihood. However, as shown in Fig. 9(b), the original data (represented by blue dots) still include some peaks during this period. Model collapse also affects the peak value generation capabilities. This is evident in Fig. 9(b), where most of the red dots (representing FCPFlow) are concentrated in a relatively lower value range compared to t-Copula and the original data. This could also explain why FCPFlow achieves a worse MAPE value than t-Copula. In contrast, in Fig. 9(a), where FCPFlow demonstrates superior performance, the *(time, peak)* points are distributed more similarly to the original data along both the x-axis (time) and $y$-axis (peak value).

In summary, while FCPFlow exhibits underperformance in the UK weather dataset, it consistently outperforms other benchmarks across the majority of experiments, reinforcing its overall efficacy.

**Table 6**
Peak generation evaluation.

| Models | GE[a] | NL | UK | AUS | USA | UK weather |
|---|---|---|---|---|---|---|
| *Euclidean distances between the (time, peak) centers* | | | | | | |
| t-Copula | **0.43** | 1.43 | 0.50 | 1.28 | 0.61 | **0.56** |
| **FCPFlow** | 0.66 | **0.25** | **0.05** | **0.07** | **0.22** | 1.23 |
| *MAPE of peaks in winter* | | | | | | |
| t-Copula | / | 23.37 | 22.80 | 11.09 | 8.71 | **20.08** |
| **FCPFlow** | / | **19.32** | **10.26** | **6.70** | **7.94** | 27.17 |
| *MAPE of Peaks in Summer* | | | | | | |
| t-Copula | / | 12.64 | 16.25 | 29.74 | 3.04 | **2.16** |
| **FCPFlow** | / | **5.01** | **1.58** | **5.06** | **2.02** | 11.66 |

[a] Season information is not available for the generated RLPs.

## 7. Simulation results for probabilistic scenario generation

### 7.1. Overall performance analysis

The developed FCPFlow model is also capable of scenario generation. In this section, we compare the FCPFlow model's performance with other generative model-based scenario generation methods, cVAE, cNICE, and cWGAN-GP. We use the NL, USA, and UK datasets in Table 2.

The NL, UK, and USA datasets are split into a test set (20% of data) and a training set (80% of data). All models are designed to take a complete RLP of the previous day as a condition and generate the scenarios of the next day. The evaluation metrics used are PL, CRPS, and MSE between the actual and average of the generated scenarios, as introduced in Section 5.3.3. We use the average PL and CRPS over time steps for comparison. We select the models with the smallest MSE during the training.

Fig. 10 illustrates the generated outcomes. From Fig. 10, we can observe that the FCPFlow model performs better in modeling critical aspects of the load, such as peaks, valleys, and volatility. Specifically, Fig. 10(a) demonstrates the FCPFlow model's proficiency in accurately generating most peaks, in contrast to cVAEA, which tends to overestimate, and cNICE, which generally has lower peak values. This pattern persists across other datasets. In the case of Fig. 10(c), which represents the most volatile scenario, although all models struggle to generate the highest peak accurately, the FCPFlow model successfully generates most of the remaining peaks and valleys.

Table 7 provides a comprehensive quantitative comparison across models, highlighting the superior performance of FCPFlow over similar models. Specifically, FCPFlow achieves the lowest MSE values on the NL and UK datasets, while on the US dataset, the MSE loss slightly lags behind cWGAN-GP. Furthermore, FCPFlow consistently exhibits significantly lower errors, ranging from 16% to 64% smaller PL errors and 5% to 46% smaller CRPS errors compared to other models. This analysis confirms the efficacy of FCPFlow not only in RLP generation but also as an advanced probabilistic scenario generation method.

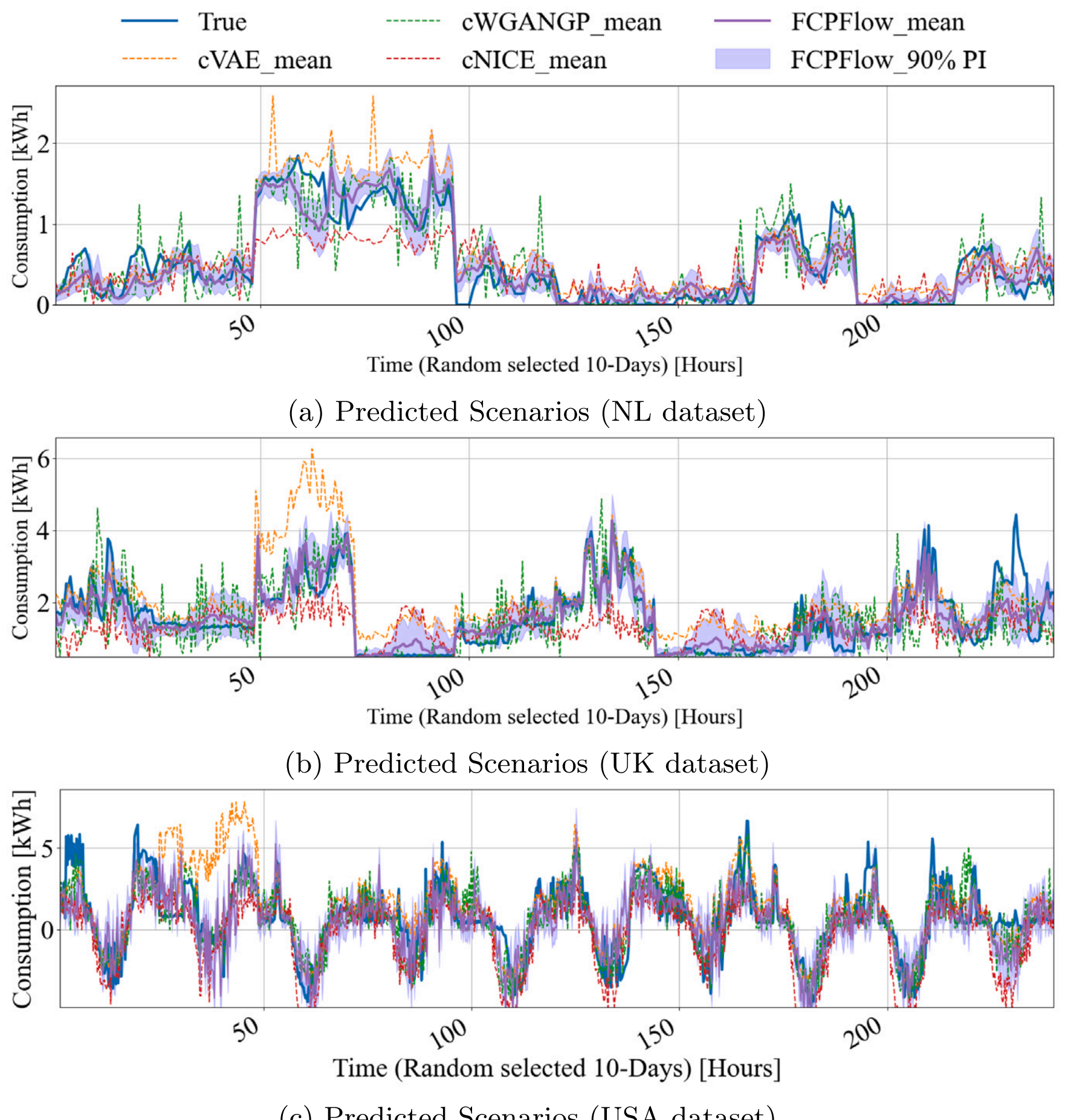

**Fig. 10.** Scenario generation for three datasets NL, UK, and USA. Different colors represent the true observation and the average of the generated scenarios. The blue area represents the 90% prediction interval of the FCPFlow model.

**Table 7**
Results of evaluation metrics for probabilistic load prediction.

| Model | PL | MSE | CRPS |
|---|---|---|---|
| *NL dataset 60 min resolution* | | | |
| cVAE | 0.1068 | 0.3616 | 0.4383 |
| cWGAN-GP | 0.1031 | 0.2302 | 0.4993 |
| cNICE | 0.1038 | 0.2406 | 0.5082 |
| **FCPFlow** | **0.0551** | **0.2052** | **0.4131** |
| *UK dataset 30 min resolution* | | | |
| cVAE | 0.0884 | 1.1491 | 1.6134 |
| cWGAN-GP | 0.1093 | 0.5511 | 2.0100 |
| cNICE | 0.1210 | 0.6913 | 2.8822 |
| **FCPFlow** | **0.0427** | **0.5412** | **1.5326** |
| *USA dataset 15 min resolution* | | | |
| cVAE | 0.2306 | 0.7838 | 1.1609 |
| cWGAN-GP | 0.2283 | **0.4865** | 1.064 |
| cNICE | 0.2846 | 0.7177 | 1.396 |
| **FCPFlow** | **0.1911** | 0.4913 | **0.8799** |

### *7.2. Empirical analysis on peaks and valleys*

In the previous section, we evaluated the overall performance of FCPFlow, demonstrating its superior performance compared to other benchmarks. However, as FCPFlow is a black-box model, and metrics like MSE, PL, and CRPS only assess overall performance, In this section, we conduct an empirical analysis on FCPFlow's performance in extreme cases by zooming in on its generated scenarios at the peaks and valleys in the dataset.

Fig. 11 presents FCPFlow's results for high peaks and low valleys with a 90% prediction interval, the upper prediction boundary (red curve), and the lower prediction boundary (blue curve). From Fig. 11, we can observe that while not all true RLPs fall within the 90% prediction intervals for all examples, the majority are covered, indicating the reliability of the predicted intervals. Additionally, there are no unreasonable predictions for the highest and lowest prediction boundaries, as no extreme deviations from the true RLPs are observed. As with other deep generative models, it is challenging to show the reliability of FCPFlow quantitatively. Therefore, we perform a case-by-case analysis to demonstrate that FCPFlow remains robust under extreme conditions.

## 8. Computational cost analysis

In this section, we discuss the computational cost of FCPFlow relative to other models, from theoretical and experimental perspectives.

While providing an exact theoretical time complexity is challenging due to model variations, we can qualitatively explain why FCPFlow is computationally more expensive than models such as VAE and WGAN-GP. Assume the input dimension is $T$ and all models are primarily composed of FNNs with Batch Normalization and ReLU activations. The per-layer time complexity for such an FNN is

$$\underbrace{O(N d_{i-1} d_i)}_{\text{Dense Mul}} + \underbrace{O(N d_i)}_{\text{BN}} + \underbrace{O(N d_i)}_{\text{Activation}} = O(N d_{i-1} d_i), \quad (38)$$

where $N$ is the batch size, and $d_{i-1}$ and $d_i$ are the input and output dimensions of layer $i$. The total time complexity scales linearly with

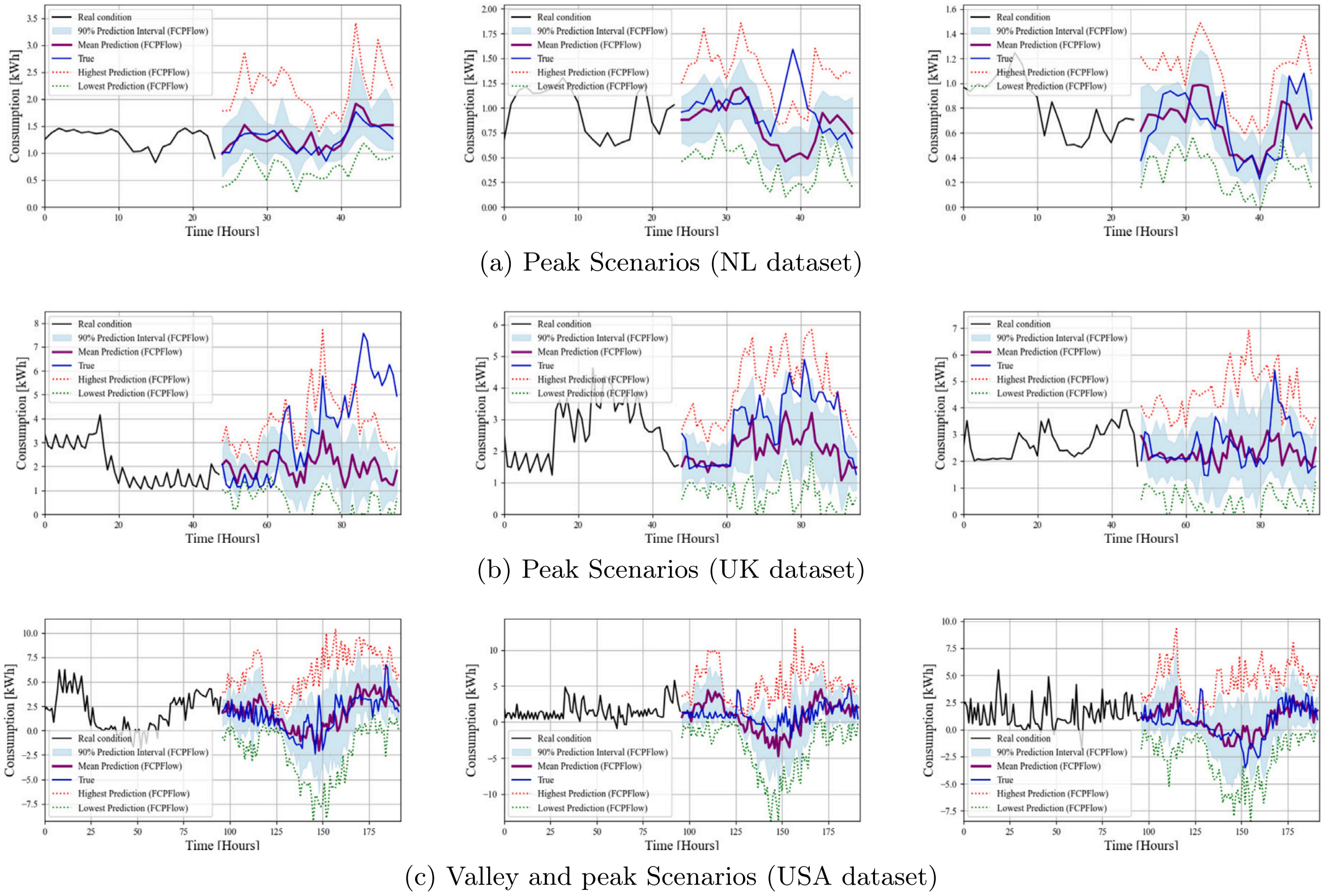

(a) Peak Scenarios (NL dataset)

(b) Peak Scenarios (UK dataset)

(c) Valley and peak Scenarios (USA dataset)

**Fig. 11.** Scenarios generation for three datasets NL, UK, and USA of high peaks and low valleys in the dataset. It can be observed that most true scenarios fall within the 90% prediction intervals. Moreover, the prediction boundaries show no extreme deviations from the true scenarios. (For interpretation of the references to color in this figure legend, the reader is referred to the web version of this article.)

the number of layers $L$ is

$$O\left(N\sum_{i=1}^{L} d_{i-1}d_i\right). \tag{39}$$

While all models are combined with FNNs, they have a similar theoretical time complexity, but the practical runtime differs. Flow-based models like NICE and FCPFlow require frequent evaluations of exponential and logarithmic functions, which are much more expensive per element than addition and multiplication. Specifically, according to computational benchmarks [50], each exponential or logarithmic operation could be 10 to 20 times more time-consuming per element than a multiplication. This means that, per operation, $\exp(\cdot)$ and $\log(\cdot)$ are slower. These additional costs, along with log-determinant computations, make flow-based models slower in practice, even though their theoretical time complexity might be similar to that of VAE and WGAN-GP.

Experimentally, given that training on large datasets is resource-intensive, and our primary interest lies in the relative computational costs among models, we selected a single household from the NL dataset, Table 2 gives detailed information of the dataset we used. We trained a small version of FCPFlow alongside other deep generative models, ensuring that all deep generative models had an approximately equivalent parameter scale of around 300,000 parameters. The Copula is implemented in GPU, while deep learning models are trained on NVIDIA A10. All models are trained for 260,000 steps.

Table 8 presents the experimental results, highlighting that FCPFlow achieves the lowest MMD value (0.0029), with an average computation time of 0.0920 s per sample. This is approximately four times more than the computational time required by VAE (0.0234 s) and nearly twice that the required by WGAN-GP. Additionally, while t-Copula is less accurate, it models data in just 1.7 s, whereas FCPFlow requires hours of training ($0.0920\times 260,000$ s). Based on the experimental results, FCPFlow is more suitable for high-quality RLP generation, while Copula may be a better choice for faster, lower-quality generation.

**Table 8**
Results of computational cost analysis.

| Model | Min MMD | Step of Min MMD | Ave time per step [s] |
|---|---|---|---|
| NICE | 0.0114 | 22,512 | 0.0638 |
| VAE | 0.2389 | 1900 | 0.0234 |
| WGAN-GP | 0.0837 | 201,216 | 0.0494 |
| **FCPFlow** | **0.0029** | 251,572 | 0.0920 |
| t-Copula | 0.1122 | *Time cost for modeling:* 1.7136 | |

**Min MMD**: The minimum MMD reached during training.
**Step of Min MMD**: The step at which Min MMD is reached.
**Ave Time per Step [s]**: The average time per training step in seconds.

Fig. 12 shows the evolution of MMD values over time and training steps. From Fig. 12 we can see that while NICE initially outperforms FCPFlow up to approximately 80,000 steps, FCPFlow ultimately achieves the lowest MMD value (0.0029) as training progresses.

## 9. Data requirement analysis

One significant application of RLP modeling is addressing the issue of data inaccessibility. However, this presents a dilemma, as any generative model requires at least some data for training. The relationship between the scale of available training data and the resulting generation performance remains an unexplored problem in RLP modeling. In this section, we aim to provide insights into this problem.

We use the NL dataset, as detailed in Table 2, with 80% of the data allocated to the training set and 20% to the test set. Instead of utilizing the entire training set, we train multiple FCPFlow models using only 10%, 30%, 60%, and 100% of the training data, subsequently evaluating the models' performance on the test set. The main purpose of this section is to gain insight into the relative performance of the models concerning the amount of available data. Therefore, given the large computational resources required for larger models, we train a set of relatively small FCPFlow models (approximately 100,000 parameters) for each data scenario and ensure that all models have the same parameter scale to eliminate the impact of model size on performance. We train the model in all scenarios with 100,000 steps.

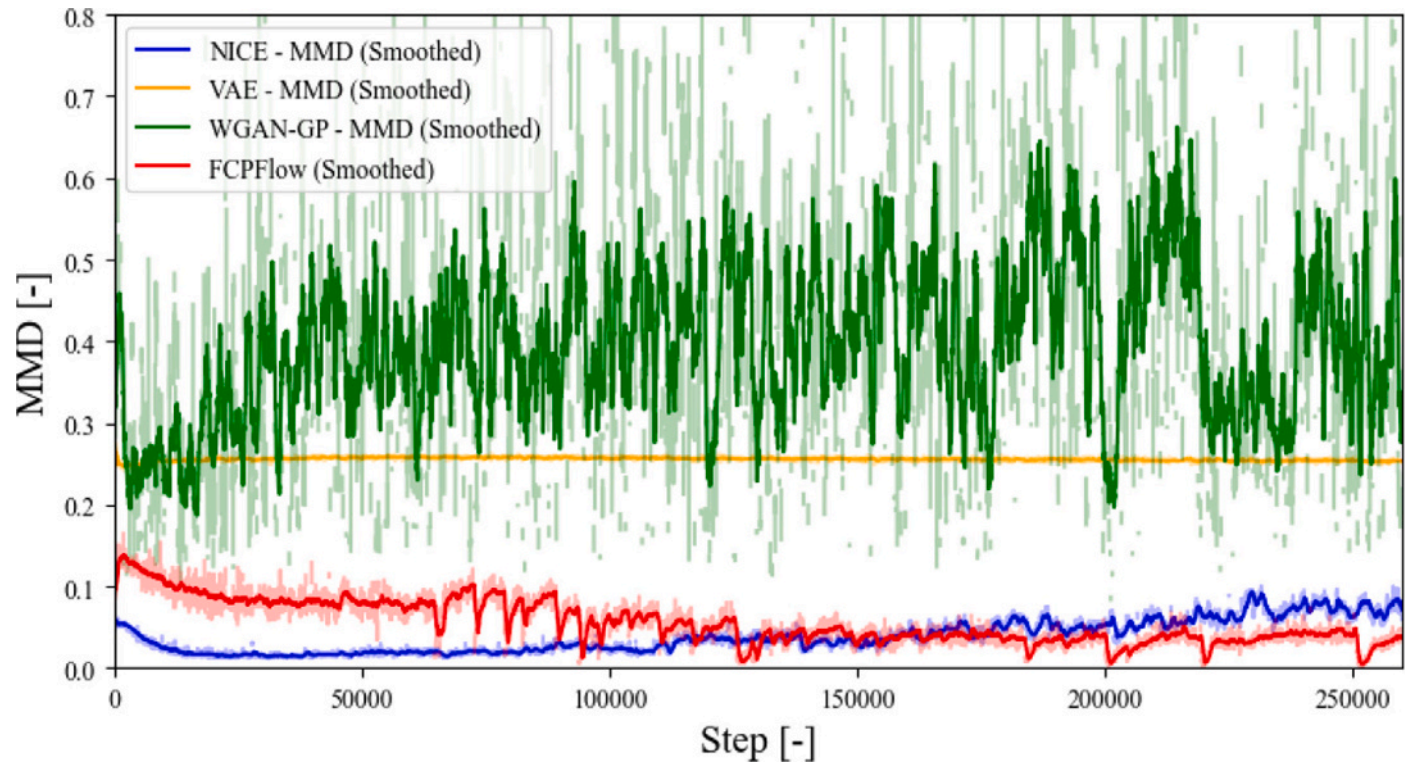

**Fig. 12.** MMD values for different models over steps. Although NICE outperforms FCPFlow initially, FCPFlow achieves the lowest MMD (0.0029) by the end.

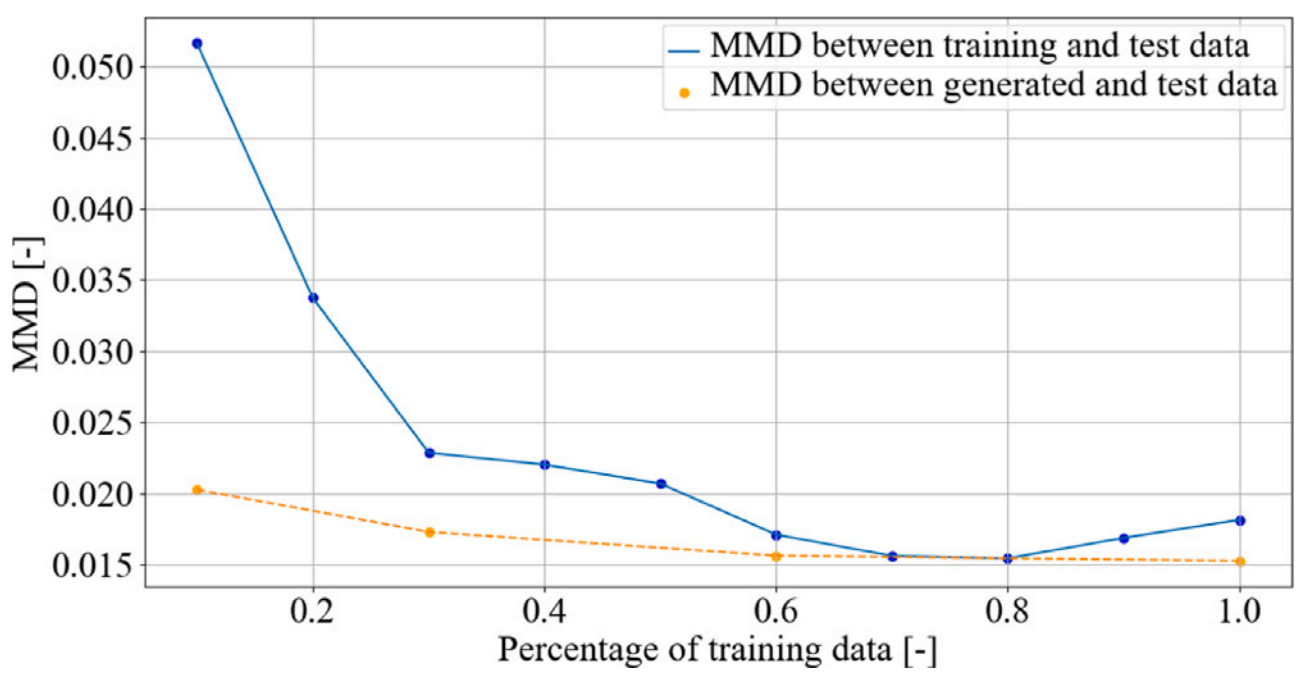

**Fig. 13.** The blue curve shows the MMD values between the test set and the training set using different percentages of data ranging from 10% to 100%. The orange curve shows the MMD values between the test set and the generated RLP data using 10%, 30%, 60%, and 100% of the training data.

The orange curve in Fig. 13 displays the MMD values between the generated RLP data (using 10%, 30%, 60%, and 100% of the training data) and the test dataset. As expected, increasing the amount of training data generally improves the model's performance. In Fig. 13, we also plot the MMD values between different scales of the training set and the test set. Similarly, as the scale of the training set increases, the MMD between the training set and the test set decreases. We believe that this trend, represented by the blue curve in Fig. 13, serves as an important reference for RLP generation. In principle, data generated from any model trained without an external dataset should generally follow this trend and adhere to this scale. The reason the MMD between the generated RLP and the test data is smaller than the MMD between the training data and the test data is that the model is selected based on the best MMD in the test set during training. If FCPFlow were trained solely using the training set without reference to the test set, its performance would more closely resemble the blue curve in Fig. 13.

## 10. Discussions

Fig. 14 summarizes the evaluation results from experiments of RLP generation and and scenario generation. The average scores for five conditional generation experiments (UK, AUS, USA, NL, UK Weather) are computed. To facilitate understanding, the smaller the area in Fig. 14, the better the model's overall performance. Based on this, the FCPFlow generally performs better than all other models. In generating RLPs, t-Copula demonstrates superior performance compared to GMMs in terms of MSE.A, primarily because it models temporal dependence while estimating the marginal distributions separately, enabling more efficient and flexible dependence modeling [28]. Deep learning models (e.g., DDPM, WGAN-GP, VAE, and FCPFlow) exhibit strong capabilities in capturing temporal correlations [35] as shown in Fig. 14(a). Among these, FCPFlow achieves the best performance in modeling temporal dependencies, as evidenced by the lowest MSE.A observed in Fig. 14(a–b). However, Deep generative models, such as WGAN-GP, often fall short of accurately reflecting the overall statistical properties, such as mean and variance. This limitation stems from the fact that models like GAN and VAE do not inherently model probability densities directly. Flow-based models address this limitation by explicitly approximating the probability density since the optimization of flow-based models responds to *Change of Variable Theorem* expressed in (4), thereby ensuring that overall statistical characteristics are better captured. Fig. 14(c) further supports this finding by demonstrating that, in scenario generation tasks, FCPFlow exhibits superior statistical performance compared to other benchmark deep generative models. Despite their strengths, traditional flow-based models have lacked the modeling capabilities seen in other deep generative models. This is because conventional flow-based models have to ensure invertibility. Therefore, flow-based models are not as flexible as other generative models. The proposed FCPFlow model retains the probabilistic precision of flow-based models while enhancing their modeling capacities for RLP data by introducing invertible linear layers and invertible normalization layers. Therefore, the FCPFlow model shows excellent performance in simultaneously capturing the temporal correlation and overall statistical characteristics of RLPs.

Copulas models offer the benefits of quick modeling and relatively robust capabilities as highlighted in Section 8. However, its assumptions constrain its performance, leading to challenges in accurately modeling complex correlations across different time steps. This limitation becomes apparent in our experiments, where Copulas' effectiveness varies, particularly with the GE and NL datasets.

One shortage of FCPFlow is the relatively long training time, especially compared with GAN and VAE as we discussed in Section 8. But we think, since we usually do not need to process billions of data in distribution systems, this shortage is acceptable in most of applications.

In Section 9, we highlighted an important relationship between the performance of generative models and the amount of available training data. Theoretically, information cannot be generated from nothing, and information passing through any channel will inevitably suffer some loss (*Second Law of Thermodynamics*). This implies that a trained model (and its generated data) may not contain more information than the training dataset itself assuming no access to the test set during training. However, augmented data may more comprehensively and efficiently represent the real data distribution by generating additional samples [51], thereby reducing information loss in subsequent channels (such as higher-level models trained on the augmented data). This could explain why augmented data enhances the performance of higher-level models.

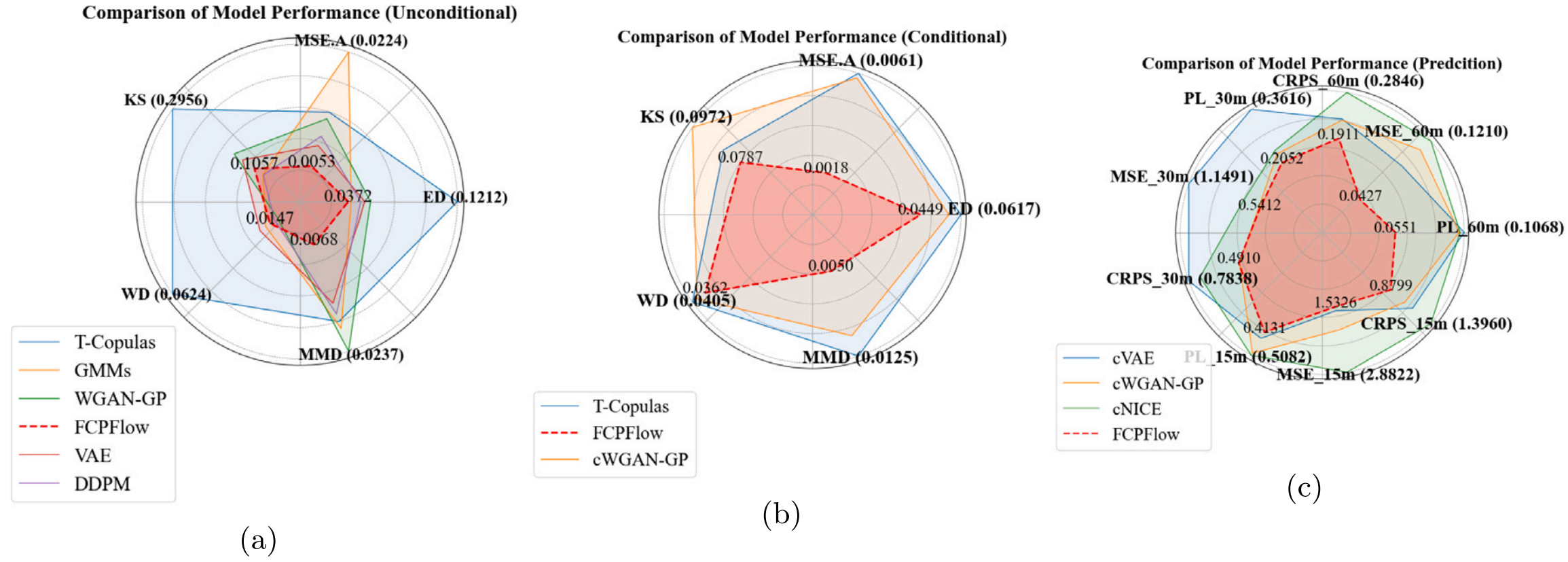

**Fig. 14.** A summary of evaluation results of all experiments. Figure (a) is for unconditional generation (GE), figure (b) is for conditional generation, and figure (c) is for load prediction. The average scores for five conditional generation experiments (UK, AUS, USA, NL, UK Weather) are computed and used for plotting. The smaller the area, the better the overall performance. Additionally, the metrics are normalized.

Moreover, we only demonstrate FCPFlow's performance in RLP generation in this paper. However, we think the potential applications of FCPFlow could extend far beyond these domains. Leveraging the *Change of Variable Theorem* in (4), a well-trained FCPFlow model can approximate the probability of specific profiles, $p_X(x)$ or $p_X(x|c)$. This could position FCPFlow as a viable tool for tasks in energy systems such as anomaly detection [52], load profile classification [53], etc.

## 11. Conclusion

This paper introduced the FCPFlow model, a novel flow-based architecture tailored to RLP generation. We conducted extensive experiments to evaluate FCPFlow's efficacy in three key areas: unconditional RLP generation (benchmarking against t-Copulas, GMMs, and WGAN-GP), conditional RLP generation (benchmarking against t-Copula and cWGAN-GP), and scenario generation (benchmarking against cNICE, cWGAN-GP, and cVAE). The FCPFlow model exhibits superior performance across all tested scenarios. Notably, the FCPFlow model combines the strengths of deep generative models, such as high stability and effectiveness in capturing temporal correlations and high-dimensional features, but also excelling in modeling overall statistical features, as evidenced by low ED, WD, and KS, among others. One limitation of FCPFlow is its relatively long training time. However, this drawback is often acceptable for most applications, as distribution systems typically do not require processing billions of data points.

## CRediT authorship contribution statement

**Weijie Xia:** Writing – review & editing, Writing – original draft, Visualization, Validation, Software, Methodology, Investigation, Formal analysis, Conceptualization. **Chenguang Wang:** Writing – review & editing, Supervision, Conceptualization. **Peter Palensky:** Resources, Project administration, Funding acquisition. **Pedro P. Vergara:** Writing – review & editing, Supervision, Resources, Project administration, Methodology.

## Declaration of competing interest

The authors declare the following financial interests/personal relationships which may be considered as potential competing interests: Weijie Xia reports writing assistance was provided by Alliander. If there are other authors, they declare that they have no known competing financial interests or personal relationships that could have appeared to influence the work reported in this paper.

## Acknowledgments

This publication is part of the project ALIGN4Energy (with project number NWA.1389.20.251) of the research programme NWA ORC 2020 which is (partly) financed by the Dutch Research Council (NWO), The Netherlands. This research utilized the Dutch National e-Infrastructure with support from the SURF Cooperative, The Netherlands (grant number: EINF-5398).

Table 9.

## Appendix

### A.1. Illustrative example: Two-dimensional transformation in fcpflow block

To improve the clarity of our proposed FCPFlow block, we provide a step-by-step numerical example for a two-dimensional input. This example demonstrates the forward transformation process from $\boldsymbol{x}$ to $\boldsymbol{z}'''$ in one FCPFlow block, including the invertible normalization layer, the invertible linear layer, and the combining coupling layer.

*Step 1: Invertible normalization layer.* Consider an input vector $\boldsymbol{x} = [x_1, x_2] = [1.0, 2.0]$. According to Eq. (22)

$$\boldsymbol{z}' = f^{-1}_{norm}(\boldsymbol{x}) = \frac{\boldsymbol{x} - \boldsymbol{\mu}}{\sqrt{\boldsymbol{\sigma}^2 + \epsilon}}. \tag{40}$$

Assume the mean vector $\boldsymbol{\mu} = [0.5, 1.5]$, standard deviation vector $\boldsymbol{\sigma} = [0.5, 0.5]$, and $\epsilon = 0$ for this example case. The normalized output is

$$\boldsymbol{z}' = \left[ \frac{1.0 - 0.5}{\sqrt{0.5^2 + \epsilon}}, \frac{2.0 - 1.5}{\sqrt{0.5^2 + \epsilon}} \right] = [1.0, 1.0]. \tag{41}$$

*Step 2: Invertible linear layer.* The invertible linear layer applies a linear transformation

$$\boldsymbol{z}'' = f^{-1}_{lin}(\boldsymbol{z}') = \boldsymbol{W}^{-1}\boldsymbol{z}'. \tag{42}$$

Assume

$$\boldsymbol{W}^{-1} = \begin{bmatrix} 2 & 0 \\ 0 & 0.5 \end{bmatrix}, \tag{43}$$

The transformed output is

$$\boldsymbol{z}'' = \begin{bmatrix} 2 & 0 \\ 0 & 0.5 \end{bmatrix} \begin{bmatrix} 1.0 \\ 1.0 \end{bmatrix} = \begin{bmatrix} 2.0 \\ 0.5 \end{bmatrix}. \tag{44}$$

**Table 9**
Structure and parameters of proposed methods and baselines.

| Experiment | Benchmark models | FCPFlow blocks | Parameters *(for deep generative models)* |
|---|---|---|---|
| Unconditional generation | | | |
| GE | t-Copula, GMM, WGAN-GP, VAE, DDPM | 3 | Around 1.0M |
| Conditional generation | | | |
| USA | | 6 | Around 5.4M |
| NL | | 6 | Around 1.2M |
| UK | t-Copula, cWGAN-GP | 6 | Around 5.0M |
| UK-weather | | 6 | Around 5.1M |
| AUS | | 6 | Around 5.0M |
| Scenario generation | | | |
| USA | | 6 | Around 470,000 |
| NL | cWGAN-GP, cNICE, cVAE | 6 | Around 230,000 |
| UK | | 6 | Around 300,000 |

*Note:* 1 M means 1 Million trainable parameters in the model.

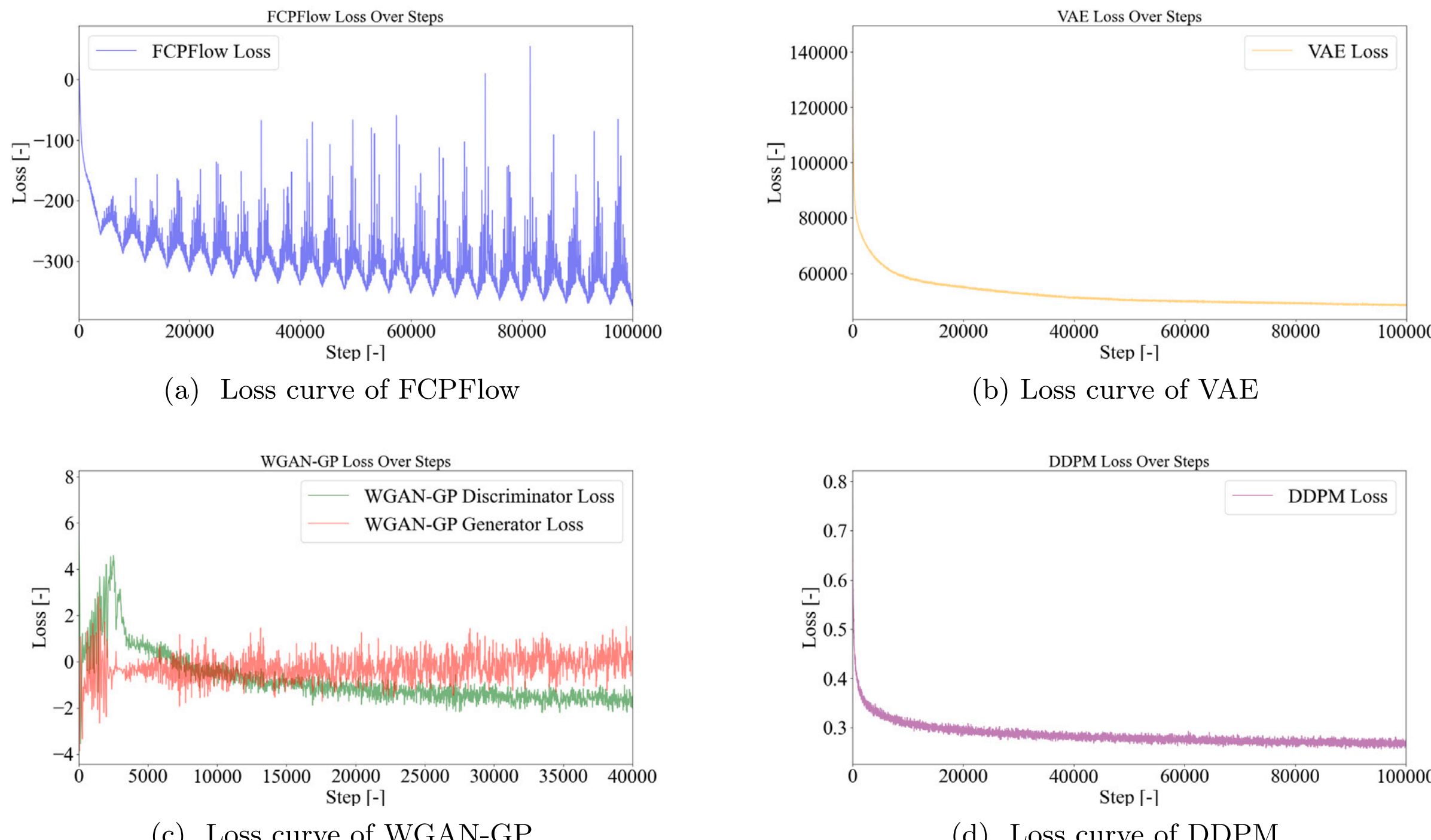

**Fig. 15.** Loss curves of different models in the unconditional generation experiment.

*Step 3: Combining coupling layer.* The combining coupling layer introduces a non-linear, invertible transformation

$$z''' = f_{ccl}^{-1}(z''). \tag{45}$$

The combining coupling layer's exact functional form is not specified here, but is designed to be invertible and commonly parameterized as affine or non-linear transformations in flow-based models. The final output after one FCPFlow block in the forward process is

$$z''' = f_{fcp}^{-1}(\boldsymbol{x}) = f_{ccl}^{-1}\left(f_{lin}\left(f_{norm}(\boldsymbol{x})\right)\right). \tag{46}$$

### A.2. Extra experimental design information

A summary of the benchmark models employed across the main experiments, along with the corresponding FCPFlow structures and the parameter scales of the deep generative models, is provided.

### A.3. Loss curves of unconditional generation experiment

Fig. 15 presents the loss curves of four deep generative models during training. All models are trained for 100,001 steps, except for WGAN-GP, which is stopped earlier at around 40,000 steps. This early stopping is applied because GAN-based models typically require more careful training procedures, and in our case, the performance of WGAN-GP began to degrade beyond around 40,000 steps.

FCPFlow, VAE, and DDPM exhibit relatively stable convergence behaviors. The oscillations observed in the loss curve of FCPFlow in Fig. 15(a) are attributed to the use of cyclical learning rates [47], as described in Section 5.4.

## Data availability

Data is shared in the Github repository.